\documentclass[10pt,twocolumn,letterpaper]{article}

\usepackage{cvpr}

\usepackage[table, dvipsnames]{xcolor}

\definecolor{cvprblue}{rgb}{0.21,0.49,0.74}

\usepackage[pagebackref,breaklinks,colorlinks,citecolor=cvprblue, urlcolor=cvprblue, linkcolor=cvprblue]{hyperref}

\def\paperID{686}
\def\confName{CVPR}
\def\confYear{2024}

\title{Move Anything with Layered Scene Diffusion}

\author{Jiawei Ren$^{1,2, *}$ \quad Mengmeng Xu$^{1}$ \quad Jui-Chieh Wu$^{1}$ \quad Ziwei Liu$^{2}$ \quad Tao Xiang$^{1}$ \quad Antoine Toisoul$^{1}$
\\[1ex]
{$^{1}$Meta AI} ~~ {$^{2}$S-Lab, Nanyang Technological University}
\\
{\tt\small \{jiawei011,ziwei.liu\}@ntu.edu.sg} \quad {\tt\small \{frostxu,jerryjcw,txiang,atoisoul\}@meta.com}
}

\newcommand\blfootnote[1]{%
\begingroup
\renewcommand\thefootnote{}{}\footnote{#1}%
\addtocounter{footnote}{-1}%
\endgroup
}

\usepackage{graphicx}
\usepackage{wrapfig}
\usepackage{makecell}
\usepackage{caption}
\usepackage{subcaption}
\usepackage{booktabs}
\usepackage{amssymb}
\usepackage{amsmath}
\usepackage{bbm}

\newcommand{\namediff}{SceneDiffusion\xspace}

\makeatletter
\DeclareRobustCommand\onedot{\futurelet\@let@token\@onedot}
\def\@onedot{\ifx\@let@token.\else.\null\fi\xspace}

\def\eg{\emph{e.g}\onedot} 
\def\ie{\emph{i.e}\onedot}

\makeatother

\usepackage{amsmath,amsfonts,bm}

\def\eqref#1{equation~\ref{#1}}

\def\1{\bm{1}}

\DeclareMathAlphabet{\mathsfit}{\encodingdefault}{\sfdefault}{m}{sl}
\SetMathAlphabet{\mathsfit}{bold}{\encodingdefault}{\sfdefault}{bx}{n}

\DeclareMathOperator*{\argmin}{arg\,min}

\begin{document}
\twocolumn[{
\renewcommand\twocolumn[1][]{#1}
\maketitle
\begin{center}
    \centering
    \vspace*{-.4cm}
    \includegraphics[width=1\textwidth]{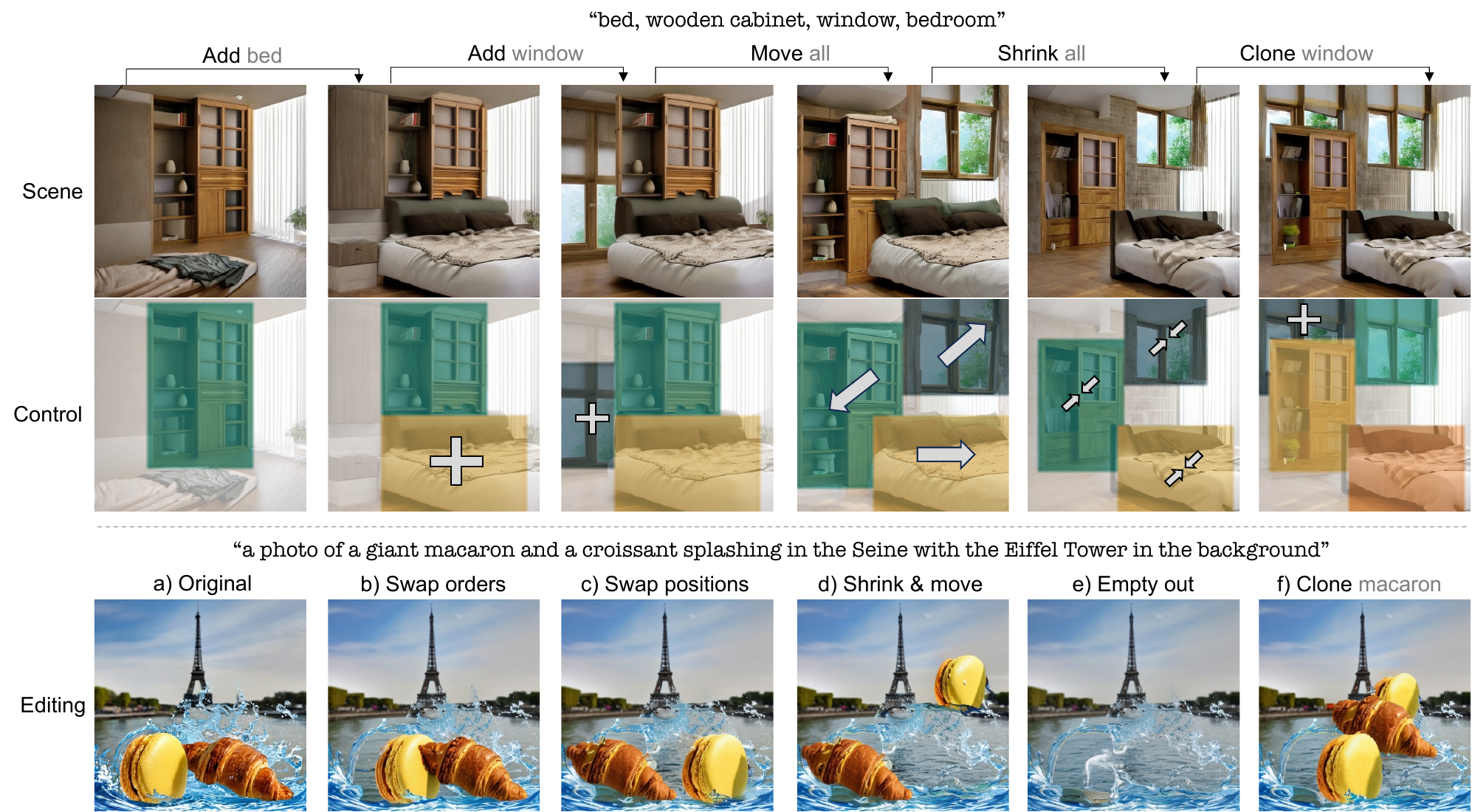}
    \captionof{figure}{\textbf{Move anything on an image.} \textbf{Top}: our approach generates \textit{playable} scenes: objects are spatially disentangled, thus can be freely moved, resized, and cloned in the scene. \textbf{Bottom}: a scene can be generated conditioned on a reference image, thus supporting extensive spatial image editing operations. Our approach is training-free and compatible with general text-to-image diffusion models. Once optimized, rendering a new layout requires \textit{less than a second} on a single GPU, allowing interactive interactions.}
\label{fig:teaser}
\end{center}
}]
\blfootnote{${^*}$Work done during an internship at Meta AI.}

\begin{abstract}
Diffusion models generate images with an unprecedented level of quality, but how can we freely rearrange image layouts? Recent works generate controllable scenes via learning spatially disentangled latent codes, but these methods do not apply to diffusion models due to their fixed forward process. In this work, we propose \textbf{SceneDiffusion} to optimize a layered scene representation during the diffusion sampling process. Our key insight is that spatial disentanglement can be obtained by jointly denoising scene renderings at different spatial layouts. Our generated scenes support a wide range of spatial editing operations, including moving, resizing, cloning, and layer-wise appearance editing operations, including object restyling and replacing. Moreover, a scene can be generated conditioned on a reference image, thus enabling object moving for in-the-wild images. Notably, this approach is training-free, compatible with general text-to-image diffusion models, and responsive in less than a second.
\end{abstract}    
\section{Introduction}
Controllable scene generation, \textit{i.e.}, the task of generating images with rearrangeable layouts, is an important topic of generative modeling~\citep{ohta1978analysis, yang2021semantic} with applications ranging from content generation and editing for social media platforms to interactive interior design and video games.

In the GAN era, latent spaces have been designed to offer a mid-level control on generated scenes~\citep{epstein2022blobgan, wang2022improving, niemeyer2021giraffe, xu2023discoscene}. Such latent spaces are optimized to provide a disentanglement between scene layout and appearance in an unsupervised manner. For instance, BlobGAN~\citep{epstein2022blobgan} uses a group of splattering blobs for 2D layout control, and GIRAFFE~\citep{niemeyer2021giraffe} uses compositional neural fields for 3D layout control. Although these methods provide good control of the scene layout, they remain limited in the quality of the generated images.
On the other hand, diffusion models have recently shown unprecedented performance at the text-to-image (T2I) generation task~\citep{sohl2015deep, ho2020denoising, dhariwal2021diffusion, rombach2021high, saharia2022photorealistic, chen2023gentron}. Still, they cannot provide fine-grained spatial control due to the lack of mid-level representations stemming from their fixed forward noising process~\citep{sohl2015deep, ho2020denoising}.

In this work, we propose a framework to bridge this gap and allow for controllable scene generation with a general pretrained T2I diffusion model. Our method, entitled \textbf{\namediff{}}, is based on the core observation that spatial-content disentanglement can be obtained during the diffusion \emph{sampling} process by denoising multiple scene layouts at each denoising step. More specifically, at each diffusion step $t$, we optimize a scene representation by first randomly sampling several scene layouts, running locally conditioned denoising on each layout in parallel, and then analytically optimizing the representation for the next diffusion step $t-1$ to minimize its distance with each of denoised result. We employ a \emph{layered} scene representation~\citep{isola2013scene, lu2020layered, kasten2021layered}, where each layer represents an object with its shape controlled by a mask and its content controlled by a text description, allowing us to compute object occlusions using depth ordering. Rendering of the layered representation is done by running a short schedule of image diffusion, which is usually completed within a second.
Overall, \namediff{} generates rearrangable scenes without requiring finetuning on paired data~\citep{zhang2023adding, mou2023t2i}, mask-specific training~\citep{rombach2021high}, or test-time optimization~\citep{poole2022dreamfusion, wang2023score}, and is agnostic to denoiser architecture designs.

In addition, to enable in-the-wild image editing,
we propose to use the sampling trajectory of the reference image as an \emph{anchor} in \namediff{}. When denoising multiple layouts simultaneously, we increase the weight of the reference layout in the noise update to keep the scene's faithfulness to the reference content. By disentangling the spatial location and visual appearance of the contents, our approach better reduces hallucinations and preserves the overall content across different editing compared to baselines~\citep{lugmayr2022repaint, epstein2023diffusion, mou2023dragondiffusion}.

To quantify the performance, we build an evaluation benchmark by creating a dataset containing 1,000 text prompts and over 5,000 images associated with image captions, local descriptions, and mask annotations. We evaluate our proposed approach on this dataset and show that it outperforms prior works on both image quality and layout consistency metrics by a clear margin on both controllable scene generation and image spatial editing tasks.

In summary, our contributions are:
\begin{itemize}
    \item We propose a novel sampling strategy, \textit{\namediff{}}, to generate layered scenes with image diffusion models.
    
    \item We show that the layered scene representation supports flexible layout rearrangements, enabling interactive scene manipulation and in-the-wild image editing.

    \item We build an evaluation benchmark and observe that our method achieves state-of-the-art performance quantitatively on both scene generation and image editing tasks.
\end{itemize}

\begin{figure*}[t!]
    \centering
    \includegraphics[width=\textwidth]{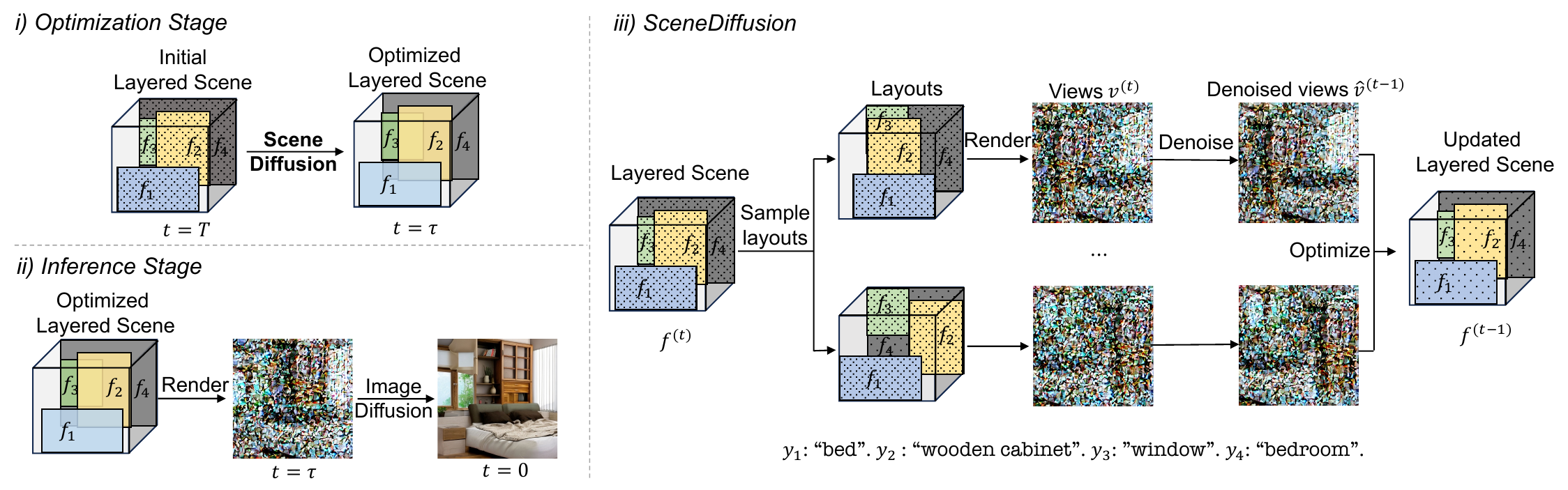}
    \caption{
    {\textbf{Method overview.} Our framework has two stages: \textbf{i) optimization stage}, we optimize a layered scene representation with \namediff{} for $T-\tau$ diffusion steps, and \textbf{ii) inference stage}, we render the optimized layered scene with $\tau$-step standard image diffusion. \textbf{iii) \namediff{}} updates the layered scene by denoising multiple randomly sampled layouts in parallel.
    In the illustration, the scene has 4 layers. Each layer consists of a feature map $f$, a mask $m$ (shown as a box), and a text prompt $y$ (shown at the bottom). At denoising step $t$, we randomly sample $N$ layouts and render them to get different views $v^{(t)}$.
    We then denoise the views using a pretrained T2I diffusion model for one step to get $\hat{v}^{(t-1)}$, which are used to update the feature maps $f^{(t)} \to f^{(t-1)}$ in the layered scene. Note that boxes here only serve as a rough geometry of objects (like blobs in ~\citet{epstein2022blobgan}), and can be replaced by more accurate masks.}
    }
    \label{fig:framework}
\end{figure*}
\section{Related Works}

\subsection{Controllable Scene Generation}
Generating controllable scenes has been an important topic in generative modeling~\citep{ohta1978analysis, yang2021semantic} and has been extensively studied in the GAN context~\citep{epstein2022blobgan, wang2022improving, niemeyer2021giraffe, xu2023discoscene}. Various approaches have been developed on applications that include controllable image generation~\citep{epstein2022blobgan, wang2022improving}, 3D-aware image generation~\citep{niemeyer2021giraffe, chan2022efficient, xu2023discoscene, hong2023evad} and controllable video generation~\citep{menapace2021playable}. Usually, control at the mid-level is obtained in an unsupervised manner by building a spatially disentangled latent space.
However, such techniques are not directly applicable to T2I diffusion models. Diffusion models employ a fixed forward process~\citep{sohl2015deep, ho2020denoising}, which constrains the flexibility of learning a spatially disentangled mid-level representation. In this work, we solve this issue by optimizing a layered scene representation during the diffusion \emph{sampling} process.
It is also noteworthy that recent works enable diffusion models to generate images grounded on given layouts~\citep{li2023gligen, zhang2023adding, mou2023t2i, gafni2022make}. However, they do not focus on spatial disentanglement and do not guarantee similar content after rearranging layouts. 

\subsection{Diffusion-based Image Editing}
Off-the-shelf T2I diffusion models can be powerful image editing tools. With the help of inversion~\citep{song2020denoising, mokady2023null} and subject-centric finetuning~\citep{ruiz2023dreambooth, gal2022image}, various approaches have been proposed to achieve image-to-image translation including concept replacement and restylization~\citep{meng2021sdedit, hertz2022prompt, tumanyan2023plug, kawar2023imagic, cong2023flatten}. However, these approaches are restricted to in-place editing, and editing the spatial location of objects has been rarely explored. Moreover, many of the approaches exploit an attention correspondence~\citep{hertz2022prompt, tumanyan2023plug, chefer2023attend, epstein2023diffusion} or a feature correspondence~\citep{tang2023emergent, shi2023dragdiffusion, mou2023dragondiffusion} with the final image, making the approach dependent to a specific denoiser architecture. 
Compared with concurrent works on spatial image editing with diffusion models using self-guidance~\citep{epstein2023diffusion, mou2023dragondiffusion} and feature tracking~\citep{shi2023dragdiffusion}, our method is different in: \emph{1)} we generate scenes that preserve the content across different spatial editing, \emph{2)} we use an explicit layered representation that gives intuitive and precise control, and \emph{3)} we render a new layout via a short schedule of image diffusion, while guidance-based approaches require a long sampling schedule and feature tracking requires gradient-based optimization for each editing.
\section{Our Approach}
\textbf{Framework Overview.} An overview of our framework is shown in \autoref{fig:framework}. In \autoref{sec:preliminary}, we briefly introduce preliminary works on diffusion models and locally conditioned diffusion. Then, in \autoref{sec:scene}, we present how we obtain a spatially disentangled layered scene with \namediff{}. Finally, in \autoref{sec:move}, we discuss how \namediff{} enables spatial editing on in-the-wild images.  

\subsection{Preliminary}\label{sec:preliminary}
\paragraph{Diffusion Models.} 
Diffusion models~\citep{sohl2015deep, ho2020denoising} are a type of generative model that learns to generate data from a random input noise. More specifically, given an image from the data distribution $x_0 \sim p(x_0)$, a \emph{fixed} forward noising process progressively adds random Gaussian noise to the data, hence creating a Markov Chain of random latent variable $x_1, x_2, ..., x_T$ following:
\begin{align}
    q(x_t|x_{t-1}) = \mathcal{N}(x_t;\sqrt{1-\beta_i}x_{t-1}, \beta_t\mathbf{I}),
\end{align}
where $\beta_1, ... \beta_T$ are constants corresponding to the noise schedule chosen so that for a high enough number of diffusion steps $x_T$ is assumed to be a standard Gaussian. We then train a denoiser $\theta$ that learns the backward process, \ie, how to remove the noise from a noisy input~\citep{ho2020denoising}. 
At inference time, we can sample an image by starting from a random standard Gaussian noise $x_T\sim\mathcal{N}(0;\mathbf{I})$ and iteratively denoise the image following the Markov Chain, i.e., by consecutively sampling $x_{t-1}$ from $p_\theta(x_{t-1}|x_t)$ until $x_0$:
\begin{equation}\label{eq:diff_sample}
    \begin{split}
    x_{t-1} = \frac{1}{\sqrt{\lambda_t}} \Bigl(x_t - \frac{1-\lambda_t}{\sqrt{1-\bar{\lambda}_t}}\epsilon_\theta(x_t, t)\Bigl) + \sigma_t\mathbf{z},
    \end{split}
\end{equation}
where $\mathbf{z} \sim \mathcal{N}(0, \mathbf{I})$, $\bar{\lambda}_t=\prod_{s=1}^t\lambda_s$, $\lambda_t=1-\beta_t$, and $\sigma_t$ is the noise scale.
\paragraph{Locally Conditioned Diffusion.} Various approaches~\citep{bar2023multidiffusion, po2023compositional} have been proposed to generate partial image content based on local text prompts using pretrained T2I diffusion models. For $K$ local prompts $\mathbf{y}=\{y_1, y_2, ..., y_K\}$ and binary non-overlapping masks $\mathbf{m}=\{m_1, m_2, ...m_K\}$, locally conditioned diffusion~\citep{po2023compositional} proposes to first predict a full image noise $\epsilon_{\theta}(x_t, t, y_k)$ for each local prompt $y_k$ with classifier-free guidance~\citep{ho2022classifier}, and then assign it to its corresponding region masked by $m_k$:
\begin{align}\label{eq:lcd}
\epsilon_{\theta}^\textrm{LCD}(x_t, t, \mathbf{y}, \mathbf{m}) = \sum_{k=1}^K m_k \odot \epsilon_{\theta}(x_t, t, y_k),
\end{align}
where $\odot$ is element-wise multiplication. 

\subsection{Controllable Scene Generation} \label{sec:scene}
Given a list of ordered object masks and their corresponding text prompts, we would like to generate a scene where object locations can be changed on the spatial dimensions while keeping the image content consistent and high quality.
We leverage a pretrained T2I diffusion model $\theta$ that generates in the image space (or latent space) $I\in\mathbb{R}^{c\times w \times h}$, where $c$ is the number of channels and $w$ and $h$ the width and height of the image, respectively. To achieve controllable scene generation, we introduce a layered scene representation in \autoref{sec:layered_represnetation} for mid-level control and propose a new sampling strategy in \autoref{sec:scene_diff}.

\subsubsection{Layered Scene Representation}\label{sec:layered_represnetation}
We decompose a controllable scene into $K$ layers $[l_k]_{k=1}^K$, ordered by the depth of the objects. 
Each layer $l_k$ has \emph{1)} a fixed object-centric binary mask $m_k\in\{0,1\}^{c\times w \times h}$ (\eg, a bounding box or segmentation mask) to show the geometric property of the object, \emph{2)} a two-element offset, $o_k\in[0;\mu_k]\times[0;\nu_k]$, indicating its spatial locations, with $\mu_k$ and $\nu_k$ defining the horizontal and vertical movement range, and \emph{3)} a feature map $f_k^{(t)}\in\mathbb{R}^{c\times w \times h}$ representing its visual appearance at diffusion step $t$.

A scene \emph{layout} is defined by the masks and their associated offsets.
The offset $o_k$ of each layer can be sampled from the movement range $[0;\mu_k]\times[0;\nu_k]$ to form a new layout.
Specially, we set the last layer $l_K$ as the background so that $m_K=\{1\}^{c\times w \times h}$ and $o_K=[0, 0]$. 
Given a layout, the layered representation can be rendered to an image, and we name the image as a \emph{view}.
Similar to prior works in controllable scene generation~\citep{epstein2022blobgan} and video editing~\citep{kasten2021layered}, we use $\alpha$-blending to composite all the layers during rendering. More concretely, the view $v^{(t)}$ can be calculated as:
\begin{equation}
    \begin{split}
    v^{(t)} &= \sum_{k=1}^K \alpha_k \odot \overline{\textrm{move}}(f_k^{(t)}, o_k), \\
    \alpha_k &= \overline{\textrm{move}}(m_k, o_k) \prod_{j=1}^{k-1}(1- \overline{\textrm{move}}(m_j, o_j)).
    \end{split}
\end{equation}
Each element in $\alpha_k\in\{0,1\}^{w\times h}$ indicates that the visibility of that location in the $k$-th latent feature map, and the function $\overline{\textrm{move}}(\cdot, o)$ means that we spatially shift the values of the feature map $f$ or mask $m$ by $o$.
The rendering process can be applied to the layered scene at any diffusion step, resulting in a view with a certain noise level. 

For initialization at diffusion step $T$, the initial feature map $f_k^{(T)}$ is independently sampled from a standard Gaussian noise  $\mathcal{N}(0, I)$ for each layer. 
It can be shown that since $\alpha$ is binary and $\sum_{k=1}^K \alpha_k^2=1$, the rendered views from the initial layered scene still follow the standard Gaussian distribution. This allows us to denoise the views directly using pretrained diffusion models.
In ~\autoref{sec:scene_diff}, we discuss how to update $f_k^{(t)}$ in a sequential denoising process.

\subsubsection{Generating Scenes with \namediff{}}\label{sec:scene_diff}
We propose \textit{\namediff{}} to optimize the feature maps in the layered scenes from Gaussian noise. 
Each \namediff{} step \emph{1)} renders multiple views from randomly sampled layouts, \emph{2)} estimates the noise from the views, and then \emph{3)} updates the feature maps.

Specifically, \namediff{}
samples $N$ groups of offset $[o_{1,n}, o_{2,n}, \cdots, o_{K,n}]_{n=1}^N$, with each offset $o_{k,n}$ being an element of the movement range $[0;\mu_k]\times[0;\nu_k]$. This leads to $N$ layout variants.
A higher number of layouts helps the denoiser locate a better mode while also increasing the computational cost, as shown in ~\autoref{sec:exp_scene_gen}.
From the $K$ latent feature maps, we
render the layouts as $N$ views $v_n\in \{v_1^{(t)}, ..., v_N^{(t)}\}$:
\begin{align}\label{eq:render}
    v_n^{(t)} = \sum_{k=1}^K \alpha_k \odot \overline{\textrm{move}}(f_k^{(t)}, o_{k,n}).
\end{align}
Then, we stack all views in each \namediff{} step and predict the noise $\{\hat\epsilon_n^{(t)}\}_{n=1}^N$ using locally conditioned diffusion~\citep{po2023compositional} described in \autoref{eq:lcd}:
\begin{align}
   \hat\epsilon_n^{(t)} = \epsilon_{\theta}^{LCD}(v_n^{(t)}, t, \mathbf{m}, \mathbf{y}), \forall n\in\{1,2,\cdots,N\}
\end{align}
where $\mathbf{m}$ are the object masks, and $\mathbf{y}$ are local text prompts for each layer. Since we can run multiple layout denoising in parallel, computing $\{\hat\epsilon_n^{(t)}\}_{n=1}^N$ brings little time overhead, while costing an additional memory consumption proportional to $N$. 
We then update the views $v_{n}^{(t)}$ from the estimated noise $\hat\epsilon_n^{(t)}$ using \autoref{eq:diff_sample} to get $\hat v_{n}^{(t-1)}$.

Since each view corresponds to a different layout and is denoised independently, conflict can happen in overlapping mask regions. Therefore, we need to optimize each feature map $f_k^{(t-1)}$ so that the rendered views from \autoref{eq:render} is close to denoised views:
\begin{equation}\label{eq:ls}
    \begin{split}
    f^{(t-1)}
     = \argmin_{f^{(t-1)}}\sum_{n=1}^N
     ||\hat{v}_n^{(t-1)}-v_n^{(t-1)}||_2^2
    \end{split}
\end{equation}
This least square problem has the following closed-form solution:
\begin{equation}\label{eq:f_update}
    \begin{split}
        f^{(t-1)}_k &= \frac{\sum_{n=1}^N \overline{\textrm{move}}(\alpha_k 
\odot \hat{v}_n^{(t-1)}, -o_{k,n})}{\sum_{n=1}^N \overline{\textrm{move}}(\alpha_k, -o_{k,n})}, \\
        & \forall k \in \{1,\cdots,K\},
    \end{split}
\end{equation}
where $\overline{\textrm{move}}(x, -o)$ denotes the values in $x$ translated in the reverse direction of $o$. The derivation for this solution is similar to the discussion in~\citet{bar2023multidiffusion}. The solution essentially sets $f^{(t-1)}_k$ to a weighted average of cropped denoised views. 

\subsubsection{Neural Rendering with Image Diffusion}\label{sec:img_diff}
We switch to vanilla image diffusion for $\tau$ steps after running \namediff{} for $T-\tau$ steps. Since the layer masks $\mathbf{m}$ like bounding boxes only serve as a rough mid-level representation instead of an accurate geometry, this image diffusion stage can be viewed as a \emph{neural renderer} that maps mid-level control to the image space~\citep{epstein2022blobgan, niemeyer2021giraffe, xu2023discoscene}. The value of $\tau$ trades off the image quality and the faithfulness to the layer mask. A value of $\tau$ in 25\% to 50\% of the total diffusion steps strikes the best balance, which usually costs less than a second using a popular 50-step DDIM scheduler~\citep{song2020denoising}. The global prompt used for the image diffusion stage can be separately set. In this work, we mainly set the global prompt to the concatenation of local prompts in the
depth order $y_\textrm{global} = <y_1, y_2, \dots, y_K>$ and find this simple strategy sufficient in most cases.

\subsubsection{Layer Appearance Editing}
The appearance of each layer can be edited individually via modifying local prompts. Objects can be restyled or replaced by changing the local prompt to a new one and then performing \namediff{} using the same feature map initialization. 

\subsection{Application to Image Editing} \label{sec:move}
\namediff{} can be conditioned on a reference image by using its sampling trajectory as an \emph{anchor}, allowing us to change the layout of an existing image. 
Concretely, when a reference image is given along with an existing layout, we set the reference image to be the optimization target at the final diffusion step, \ie, an anchor view denoted as $\hat{v}_a^{(0)}$. Then, we add Gaussian noise to this view at different diffusion noise levels, creating a trajectory of anchor views at different denoising steps.
\begin{equation}
    \begin{split}
   \hat{v}_a^{(t)} = \sqrt{1-\beta_t}\hat{v}_a^{(0)} + \beta_t\epsilon, \; \forall t \in [1,\cdots,T],
    \end{split}
\end{equation}
where $\epsilon\sim \mathcal{N}(0,1)$. In each diffusion step, we use the corresponding anchor view $\hat{v}_a^{(t)}$ to further constraint $f^{(t-1)}$, which leads to an extra weighted term in \autoref{eq:ls}:
\begin{equation}\label{eq:ls2}
    \begin{split}
      f^{(t-1)} 
     &=\argmin_{f_.^{(t-1)}} \sum_{n}
     w_n ||\hat{v}_n^{(t-1)}-v_n^{(t-1)}||_2^2\\
     w_n &=
    \begin{cases}
      w & \text{if} \; n = a, \\
      1 & \text{otherwise.}
    \end{cases}
    \end{split}
\end{equation}
where $n\in\{1,\cdots,N\}\cup\{a\}$, and $w$ controls the importance of $\hat{v}_a^{(t)}$. A large enough $w$ produces good faithfulness to the reference image, we set $w=10^{4}$ in this work. The closed-form solution of this equation is similar to \autoref{eq:f_update} and can be found in supplementary material.
\section{Experiments}
\begin{figure*}[t!]
    \centering
    \includegraphics[width=0.95\textwidth]{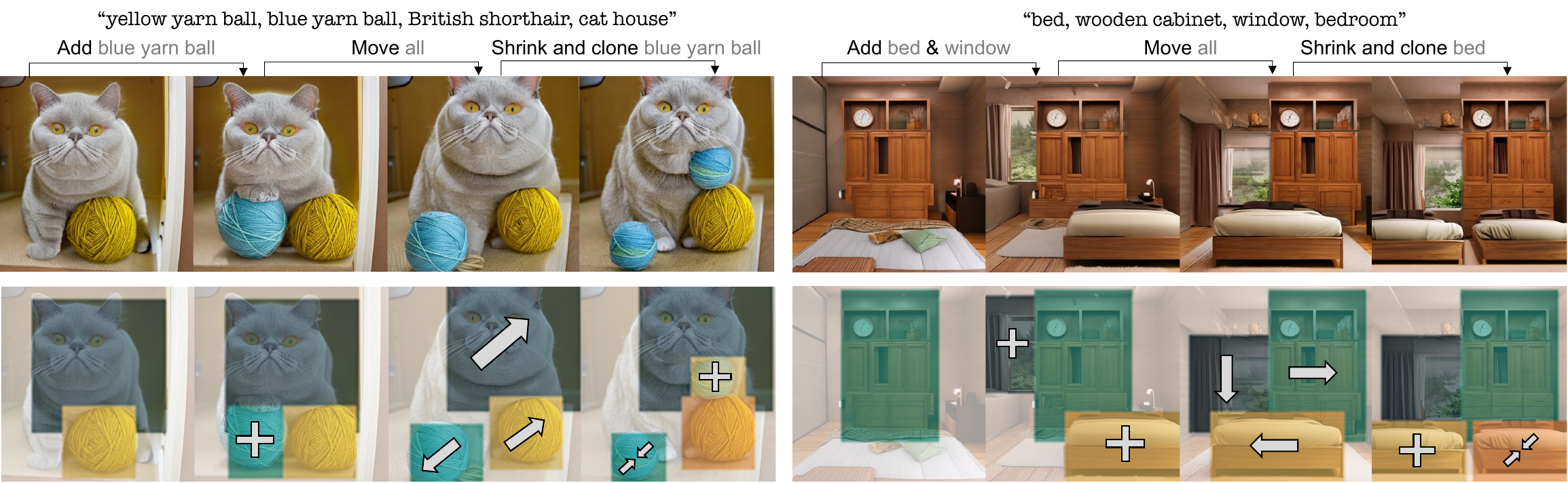}
    \caption{
    {\textbf{Sequential manipulations.} Our generated scenes can be manipulated by operating on layers sequentially.}
    }
    \label{fig:sequential_manipulations}
\end{figure*}
\begin{figure*}[t!]
    \centering
    \includegraphics[width=0.95\textwidth]{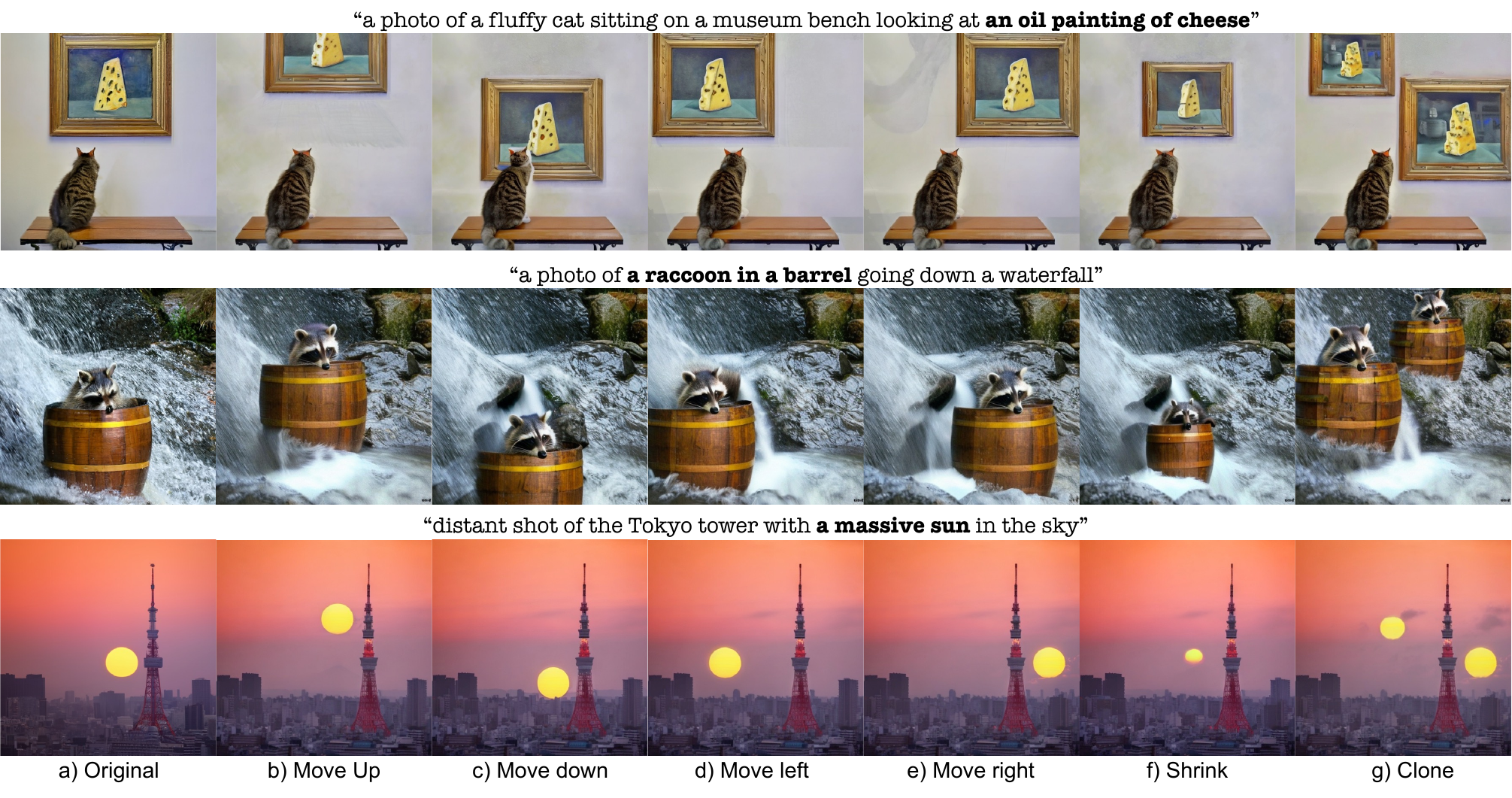}
    \caption{
    {\textbf{Object moving.} Our approach can be employed to move objects on a given image. Edited objects are shown in bold in the prompts. Examples are borrowed from \citet{epstein2023diffusion} and no access to the initial latent noise is assumed. All layouts for each example are generated from the same scene. As a result, our approach keeps the overall content consistent across different editings, which most prior works fail to achieve. A full comparison with prior works can be found in appendix.}
    }
    \label{fig:object_moving}
\end{figure*}
\begin{figure*}[t!]
    \centering
    \includegraphics[width=\textwidth]{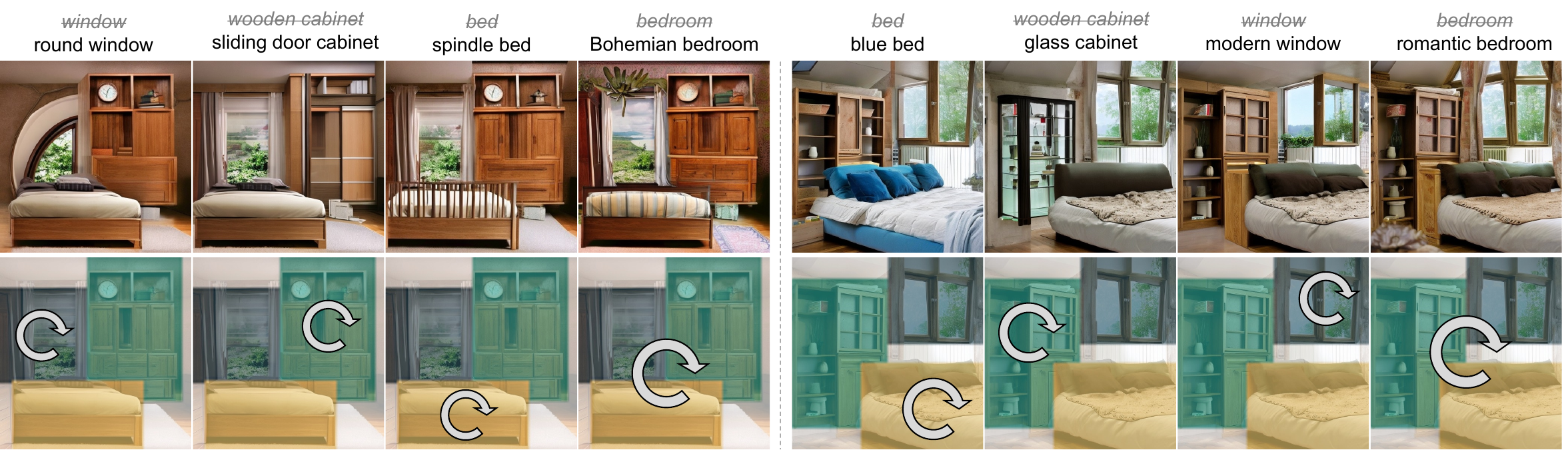}
    \caption{
    {\textbf{Restyling objects.} Adding style description to the layer prompt restyles the object when fixing the initial noise. The circular arrow shows the restyled object.}
    }
    \label{fig:restyling_object}
\end{figure*}
\begin{figure}[t!]
    \centering
    \includegraphics[width=\columnwidth]{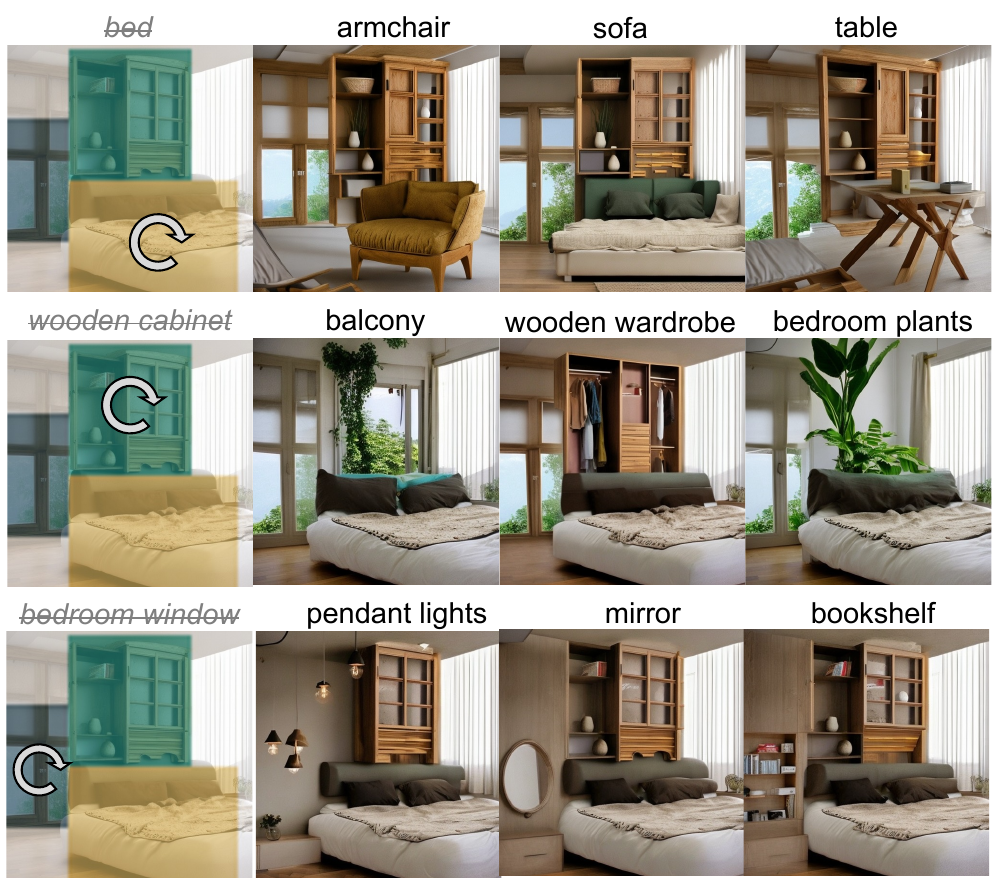}
    \caption{
    {\textbf{Replacing objects.} Objects can be changed to different objects by modifying their layer prompts without affecting other objects in the scene. The circular arrow shows the replaced object.}
     }
    \label{fig:replacing_object}
\end{figure}
\begin{figure}[t!]
    \centering
    \includegraphics[width=0.95\columnwidth]{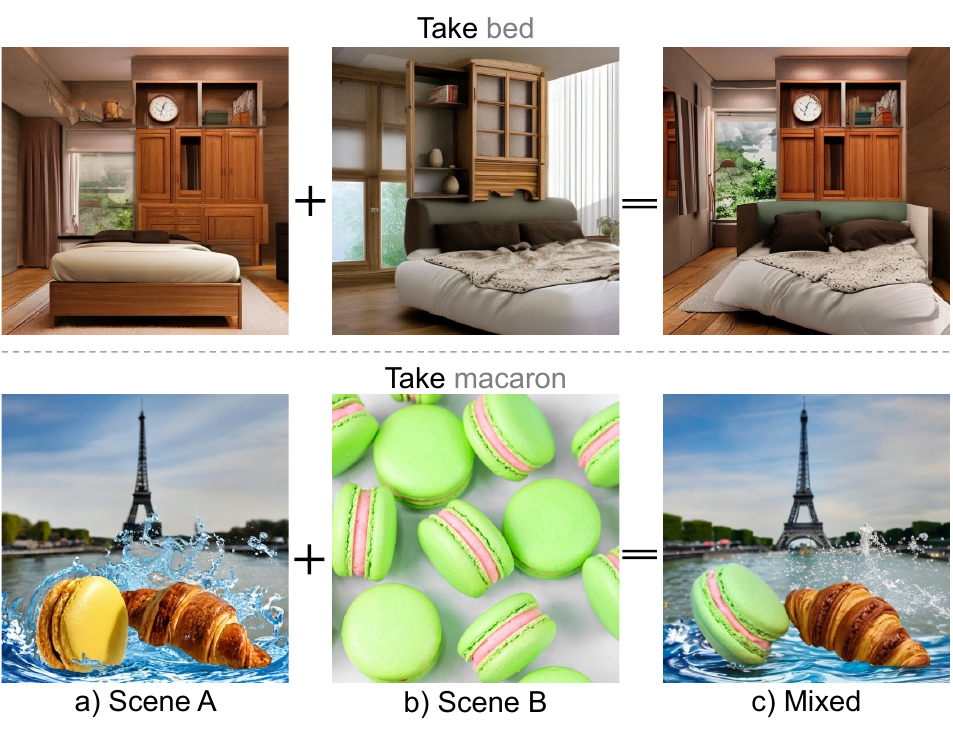}
    \caption{
    {\textbf{Mixing scenes.} One may mix scenes by copying a layer from one scene and pasting it in another scene. }
    }
    \label{fig:mixing_scenes}
\end{figure}
\begin{figure*}[h!]
    \centering
    \includegraphics[width=0.85\textwidth]{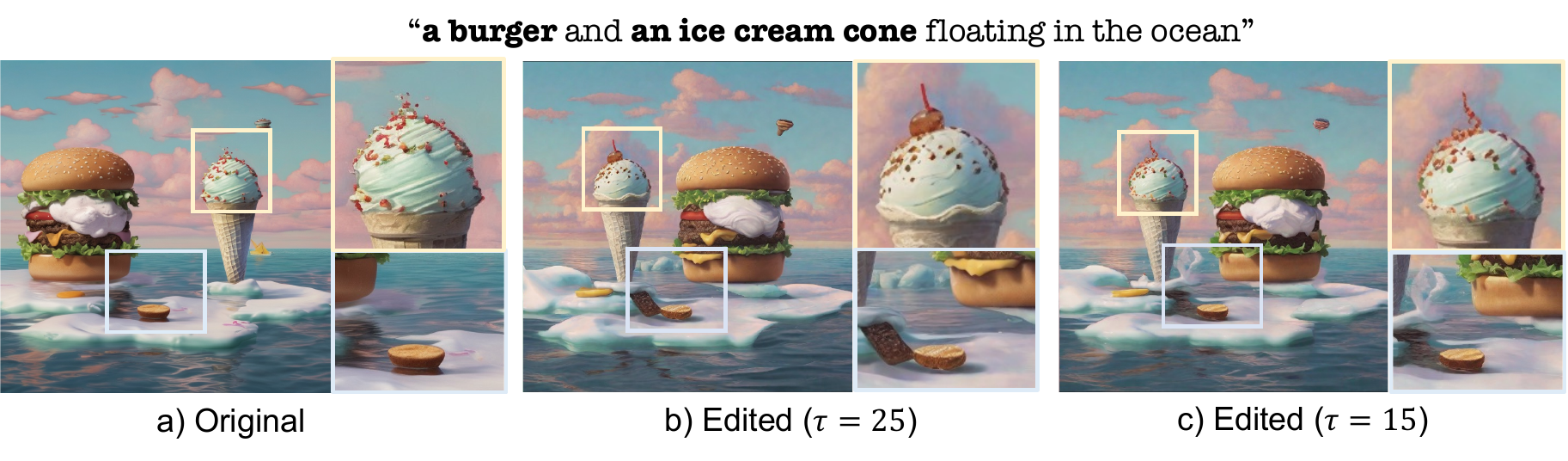}
    \caption{\textbf{Ablation on $\tau$.} We swap the locations of the two objects. Stopping \namediff{} at a later step improves consistency and prevents hallucination.}
    \label{fig:move_tradeoff}
\end{figure*}

\subsection{Experimental Setup}
We evaluate our method both \emph{qualitatively} and \emph{quantitatively}. For quantitative study, a thousand-scale dataset is required to effectively measure metrics like \emph{FID}. However, populating semantically meaningful spatial editing pairs for multi-object scenes is challenging, particularly when inter-object occlusions should be considered. Therefore, we restrict quantitative experiments to single-object scenes. Please refer to qualitative results for multi-object scenes.

\paragraph{Dataset.} We curate a dataset of high-quality, subject-centric images associated with image captions and local descriptions. Object masks are also annotated automatically using GroundedSAM~\citep{ren2024grounded}. We first generate 20,000 images from 1,000 image captions and then apply a rule-based filter to remove low-quality images, which results in 5,092 images in total. Object masks and local descriptions are then automatically annotated.

\paragraph{Metrics.}
Our main metrics for controllable scene generation are \textit{Mask IoU}, \textit{Consistency}, \textit{Visual Consistency}, \textit{LPIPS}, and \textit{SSIM}. \textit{Mask IoU} measures the alignment between the target layout and the generated image. Other metrics compare multiple generated views in the same scene and evaluate their similarity: \textit{Consistency} for mask consistency, \textit{Visual Consistency} for foreground appearance consistency, \textit{LPIPS} for perceptual, and \textit{SSIM} for structural changes. Moreover, in the image editing experiment, we report \textit{FID} to measure the similarity of the edited images to the original ones for image quality quantification.

\paragraph{Implementation} By default we set $N=8$ in our experiments. For quantitative studies, all experiments are averaged on 5 random seeds. Please refer to our supplemental document for more information on our dataset construction, metrics selection, standard deviations of experiments and implementation details.

\subsection{Controllable Scene Generation}\label{sec:exp_scene_gen}
\begin{figure}[ht]
    \centering
    \includegraphics[width=\columnwidth]{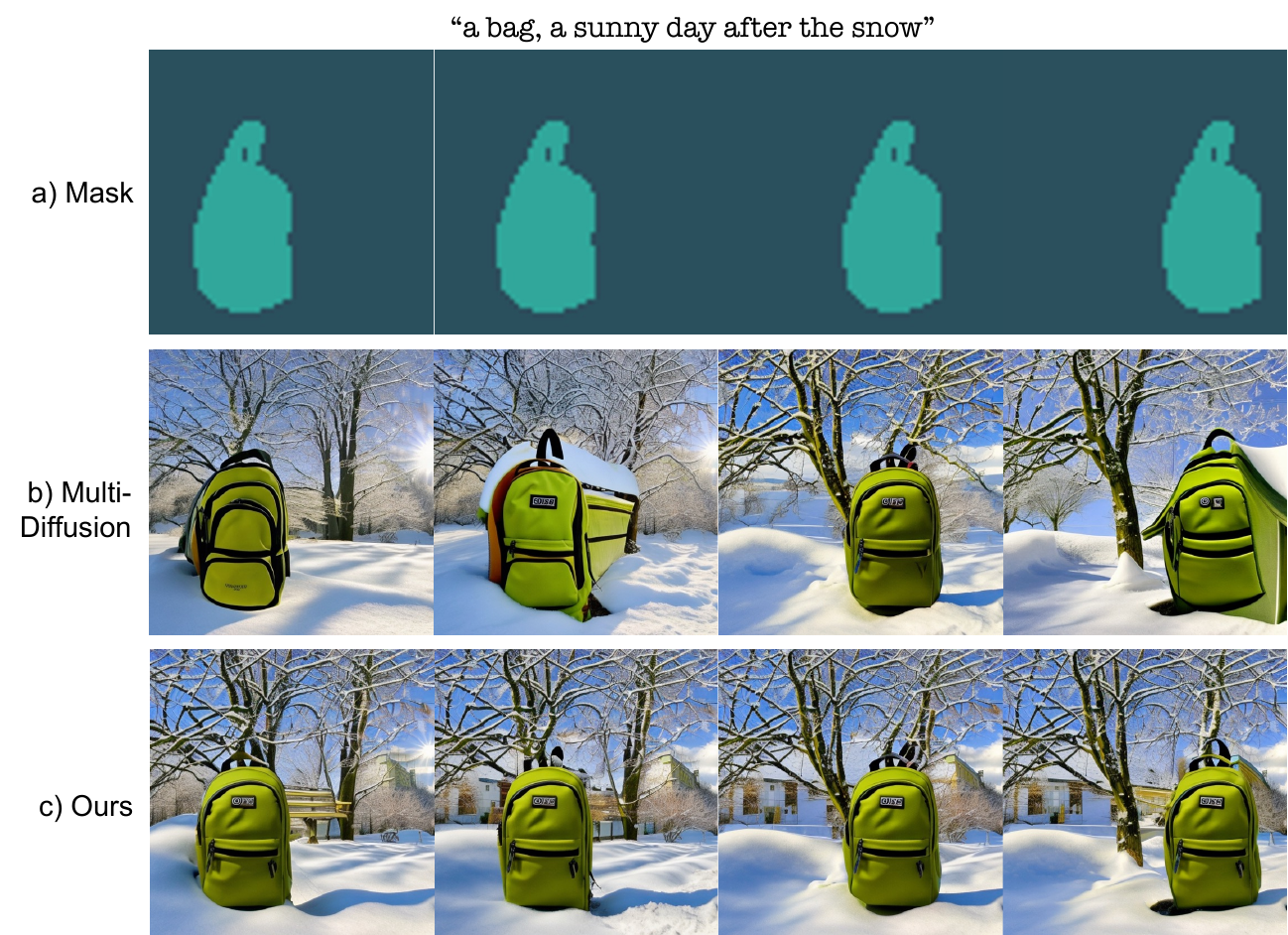}
    \captionof{figure}{\textbf{Qualitative evaluation of controllable scene generation.} Multidiffusion~\citep{bar2023multidiffusion} is able to generate a backpack in accordance to the target mask, but both the background and the object change at different layouts. Our method can produce coherent and consistent images with minimal visual appearance difference. 
    }
    \label{fig:qualitative_scene}
\end{figure}

\paragraph{Setting.} We randomly place an object mask at different positions to form random target layouts. Images should be generated conditioned on the target layouts and local prompts, and the content is expected to be consistent in different layouts. The object masks are from the aforementioned curated dataset. To reduce the chance that objects move out of the canvas, we restrict the maks position to a square centered at the original position with its side length of 40\% of the image width.  A visual example can be found in \autoref{fig:qualitative_scene}.

\paragraph{Baselines.} We compare our approach to MultiDiffusion~\citep{bar2023multidiffusion}, which is a training-free approach that generates images conditioned on masks and local descriptions. We use a 20\% solid color bootstrapping strategy following their protocol. Foreground and background noise are fixed in the same scene for better consistency.

\paragraph{Results.}
We present quantitative results in \autoref{tab:scene_comparison}, which show that \namediff{} outperforms MultiDiffusion on all metrics. For qualitative study, we show the results of sequentially manipulation our generated scenes in \autoref{fig:sequential_manipulations}.

\begin{table}[t]
\scriptsize
\vspace{-10pt}
\caption{\textbf{Quantitative comparison for controllable scene generation.} $\dagger$: without the solid color bootstrapping strategy.}
\vspace{-20pt}
\begin{center}
\begin{tabular}{lccccc}
\toprule
Method & M. IoU $\uparrow$ &  Cons.$\uparrow$ & V. Cons.$\downarrow$  & LPIPS $\downarrow$ & SSIM $\uparrow$\\
\midrule
MultiDiff.~\citep{bar2023multidiffusion}$^\dagger$ & 0.263 & 0.257 & - & 0.521 & 0.450 \\
MultiDiff.~\citep{bar2023multidiffusion} & 0.466  & 0.436 & 0.236 & 0.519 & 0.471\\
\midrule
Ours$^\dagger$ & 0.310 & 0.609  & - & \textbf{0.198} & 0.761 \\
Ours &  \textbf{0.522} &  \textbf{0.721} & \textbf{0.112} & 0.215 & \textbf{0.762}\\
\bottomrule
\end{tabular}
\label{tab:scene_comparison}
\vspace{-20pt}
\end{center}
\end{table}

\subsection{Object Moving for Image Editing }
\paragraph{Setting.} Given a reference image, an object mask, and a random target position, the goal is to generate an image where the object has moved to the target position while keeping the rest of the content similar. The aforementioned range is used to prevent moving the object out of the canvas.

\begin{table}[t]
\vspace{-10pt}
\caption{\textbf{Quantitative comparison for object moving.} $\dagger$: specialized inpainting model trained with masking.}
\scriptsize
\vspace{-20pt}
\begin{center}
\begin{tabular}{lcccccc}
\toprule
Method & FID $\downarrow$ & M. IoU $\uparrow$ & V. Cons. $\downarrow$  & LPIPS $\downarrow$ & SSIM $\uparrow$\\
\midrule
RePaint~\citep{lugmayr2022repaint} & 10.267 & 0.620 & 0.166 & 0.278 & 0.671 \\
Inpainting$^\dagger$ & 6.383 & 0.747 & 0.112 & 0.264  & 0.680 \\
\midrule
Ours & \textbf{5.289} & \textbf{0.817} & \textbf{0.075} & \textbf{0.263} & \textbf{0.709} \\
\bottomrule
\end{tabular}
\label{tab:move_comparison}
\vspace{-20pt}
\end{center}
\end{table}

\paragraph{Baselines.} We compare with inpainting-based approaches. We first crop the object from the reference image, paste it to the target location, and then inpaint the blank areas. We dilate the edge of objects for 30 pixels to better blend the object with the background. We compare our approach with two inpainting models: a standard T2I diffusion model using the RePaint technique~\citep{lugmayr2022repaint}, and a specialized inpainting model trained with masking. We set all local layer prompts in our approach to the global image caption for a fair comparison.

\paragraph{Results.}
We report quantitative results in \autoref{tab:move_comparison}. Our approach outperforms both inpainting-based baselines by a clear margin on all metrics. Qualitative results of object moving are shown in \autoref{fig:object_moving}.

\subsection{Layer Appearance Editing}
We show the results of object restyling in \autoref{fig:restyling_object} and object replacement in \autoref{fig:replacing_object}. We observe that changes are mostly isolated to the selected layer, while other layers slightly adapt to make the scene more natural. Furthermore, layer appearance can be transferred across scenes by directly copying a layer from one scene to another, as shown in \autoref{fig:mixing_scenes}. 

\begin{table}[ht!]
\scriptsize
\vspace{-10pt}
\caption{\textbf{Component analysis.}}

\vspace{-20pt}
\begin{center}
\setlength{\tabcolsep}{2pt}
\begin{tabular}{lcccccc}
\toprule
Method & \textbf{CLIP-a} $\uparrow$ & \textbf{VC} $\downarrow$ & M. IoU $\uparrow$ & Cons.$\uparrow$ & LPIPS $\downarrow$ & SSIM $\uparrow$\\
\midrule
Ours ($N$=8, $\tau$=13) & 6.12 & 0.11 & 0.51 & 0.72 & 0.22 & 0.74 \\
\midrule
w/o multiple layouts & \cellcolor{red!12.5}6.05 & \cellcolor{red!12.5}0.23 & \cellcolor{red!12.5}0.46 & \cellcolor{red!12.5}0.43 & \cellcolor{red!12.5}0.51 & \cellcolor{red!12.5}0.47 \\
w/o random sampling & \cellcolor{red!12.5}5.98 & \cellcolor{red!12.5}0.12 & \cellcolor{red!12.5}0.50 & \cellcolor{red!12.5}0.68 & 0.22 & \cellcolor{blue!12.5}0.75 \\
w/o image diffusion & \cellcolor{red!12.5}5.96 & \cellcolor{blue!12.5}0.09 & 0.51 & 0.72 & \cellcolor{blue!12.5}0.21 & \cellcolor{blue!12.5}0.76 \\
\bottomrule
\end{tabular}
\label{tab:rebuttal_scene_ablation}
\end{center}
\end{table}
\begin{table}[ht!]
\scriptsize
\vspace{-10pt}
\caption{\textbf{Analysis on $N$ and $\tau$} }
\vspace{-20pt}
\begin{center}
\setlength{\tabcolsep}{2pt}
\begin{tabular}{lc|cc|ccccc}
\toprule
$N$ & $\tau$ & Optim.$\downarrow$ & Infer.$\downarrow$ & \textbf{CLIP-a} $\uparrow$ & M. IoU $\uparrow$ &  Cons.$\uparrow$ & LPIPS $\downarrow$ & SSIM $\uparrow$\\
\midrule
8 & 13 &  17.3s & 0.82s & 6.12 & 0.514 & 0.721 & 0.224 & 0.749\\
\midrule
\cellcolor{yellow!12.5}4 & 13 &  \cellcolor{blue!12.5}9.65s & 0.82s & \cellcolor{red!12.5}5.99 & \cellcolor{red!12.5}0.491 & \cellcolor{red!12.5}0.689 & \cellcolor{red!12.5}0.225 & \cellcolor{red!12.5}0.747  \\
\cellcolor{yellow!12.5}2 & 13 &  \cellcolor{blue!12.5}5.73s & 0.82s & \cellcolor{red!12.5}5.97 & \cellcolor{red!12.5}0.481 & \cellcolor{red!12.5}0.672 & \cellcolor{red!12.5}0.229 & \cellcolor{red!12.5}0.735 \\
\midrule
8 &\cellcolor{yellow!12.5}25 & \cellcolor{blue!12.5}12.0s & \cellcolor{red!12.5}1.53s & \cellcolor{blue!12.5}6.13 & \cellcolor{red!12.5}0.502 & \cellcolor{red!12.5}0.643 & \cellcolor{red!12.5}0.276 & \cellcolor{red!12.5}0.685 \\
8 & \cellcolor{yellow!12.5}0 &  \cellcolor{red!12.5}22.9s & \cellcolor{blue!12.5}0.0s & \cellcolor{red!12.5}5.96 & \cellcolor{blue!12.5}0.515 & \cellcolor{blue!12.5}0.723 & \cellcolor{blue!12.5}0.211 & \cellcolor{blue!12.5}0.767 \\
\bottomrule
\end{tabular}
\label{tab:rebuttal_scene_tradeoff}
\vspace{-20pt}
\end{center}
\end{table}
\subsection{Ablation study}
In \autoref{tab:rebuttal_scene_ablation}, we ablate all components. We additionally measure \emph{CLIP-aesthetic} (CLIP-a) following~\cite{bar2023multidiffusion} to quantify the image quality. Without jointly denoising multiple layouts, all metrics drop drastically. With a deterministic sampling of layouts, the image quality degrades. Without the image diffusion stage, although consistency metrics slightly improve, image quality significantly deteriorates. 
In \autoref{tab:rebuttal_scene_tradeoff}, we analyze the effect of the number of views and image diffusion steps. We observe that having more views and more \namediff{} steps leads to a better disentanglement between the object and the background, as indicated by higher Mask IoU and Consistency. A qualitative comparison can be found in \autoref{fig:move_tradeoff}. We also present the accuracy-speed trade-off when limiting to a single 32GB GPU. Larger $N$ increases the optimization time. Larger $\tau$ increases the inference time.
For all ablation experiments, we use a randomly selected 10\% subset for easier implementation.
\section{Conclusion}
We proposed \namediff{} that achieves controllable scene generation using image diffusion models. \namediff{} optimizes a layered scene representation during the diffusion sampling process. Thanks to the layered representation, spatial and appearance information are disentangled which allows extensive spatial editing operations. Leveraging the sampling trajectory of a reference image as an anchor, \namediff{} can move objects on in-the-wild images. Compared to baselines, our approach achieves better generation quality, cross-layout consistency, and running speed. 
{\noindent\textbf{Limitations}.
The object's appearance may not fit tightly to the mask in the final rendered image. Besides, our approach requires a large amount of memory to simultaneously denoise multiple layouts, restricting the applications in resource-limited user cases.}
\noindent\textbf{Acknowledgments}.
This study is supported by the National Research Foundation, Singapore under its AI Singapore Programme (AISG Award No: AISG2-PhD-2021-08-018), the Ministry of Education, Singapore, under its MOE AcRF Tier 2 (MOET2EP20221- 0012), NTU NAP, and under the RIE2020 Industry Alignment Fund – Industry Collaboration Projects (IAF-ICP) Funding Initiative.
\clearpage
{
    \small
    \bibliographystyle{ieeenat_fullname}
    \bibliography{ref}
}

\clearpage
\appendix
\tableofcontents

\section{Solution to Equation 10}
The analytical solution to Equation 10 is:
\begin{equation}\label{eq:f_updat2}
    \begin{split}
   f^{(t-1)}_k &= \frac{\sum_n w_n\overline{\textrm{move}}(\alpha_k 
\odot \hat{v}_n^{(t-1)}, -o_{k,n})}{\sum_n w_n \overline{\textrm{move}}( \alpha_k,-o_{i,n})}; 
\\
\quad \forall k &\in \{1,\cdots,K\},
    \end{split}
\end{equation}
where $n\in\{1,\cdots,N\}\cup\{a\}$, $o_{k,a}$ is the layout of the given image.

\section{Discussion on Layer Masks}
\begin{figure*}[t!]
    \centering
    \includegraphics[width=\textwidth]{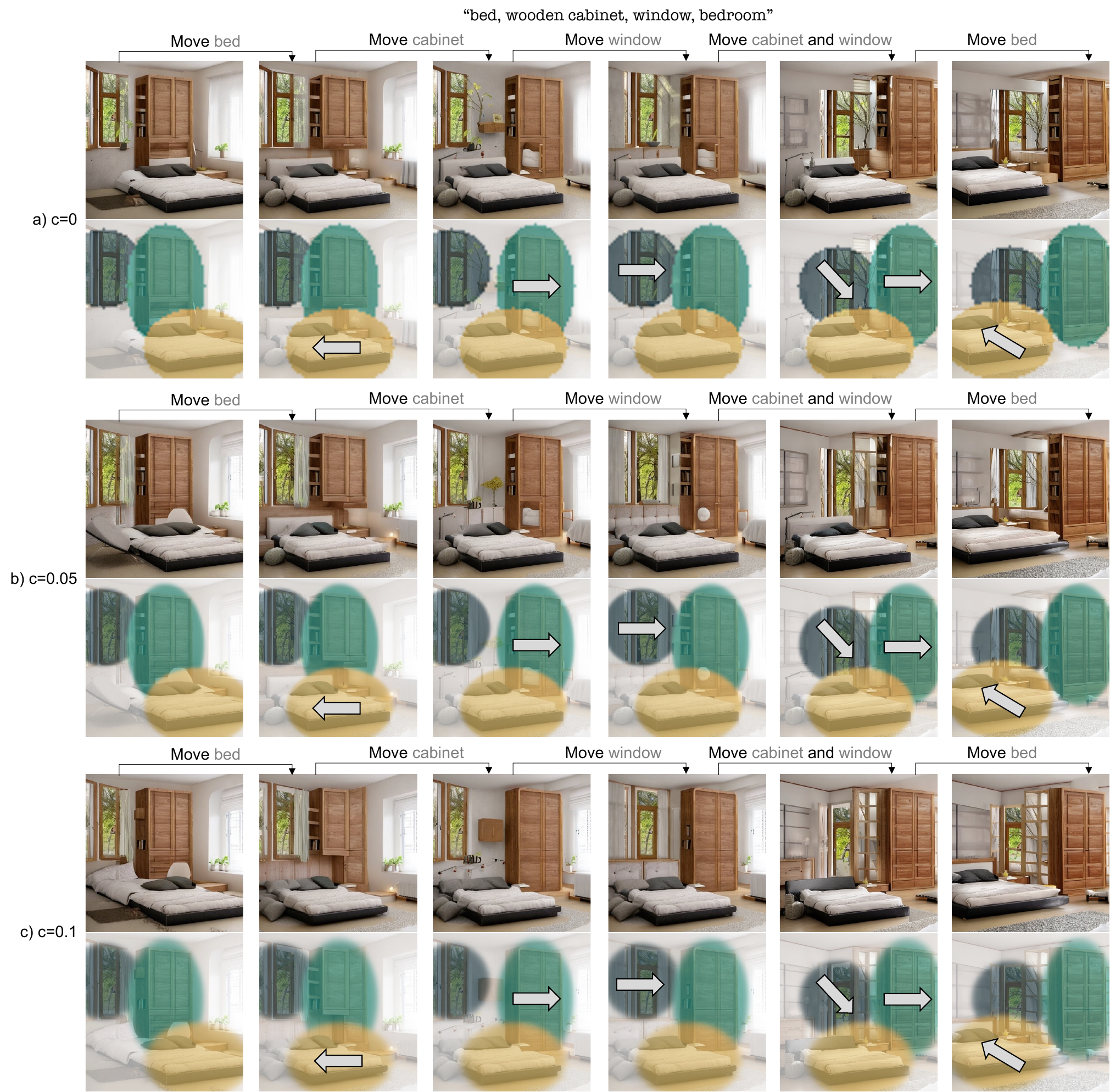}
    \caption{
    {\textbf{Blobs as layer masks.} Layer masks can also be represented using elliptical blobs instead of bounding boxes. In addition, the updated $\alpha$-blending can handle soft masks instead of binary masks.}
    }
    \label{fig:blob}
\end{figure*}
\subsection{Elliptical blob masks}
We mainly use bounding boxes for layer masks in the main paper. The layer masks can also be represented by other shapes, for example, elliptical blobs~\citep{epstein2022blobgan}. Blobs are parameterized by centroids, scales, and angles. Moreover, blobs have alpha values decaying from the centroids to soften the edges. The edge sharpness can be controlled by a parameter $c$: a smaller $c$ leads to stronger edge sharpness and $c=0$ corresponds to hard thresholding. Due to the standard Gaussian noise assumption at the initial stage of diffusion, we set $c=0$ so that alpha values are binary. We show results of using blobs for layer masks in \autoref{fig:blob}.
\subsection{Soft masks with modified $\alpha$-blending}
Soft masks can be enabled by a modified rendering equation. As discussed in the main paper, the standard Gaussian noise assumption introduced by image diffusion models requires $\sum_{k=1}^K \alpha_k^2=1$. On the other hand, the standard $\alpha$-blending described in Equation 4 results in alpha values that sum to one. Therefore, the assumption can only be fulfilled when $\alpha$ is binary. To use soft masks, we may modify $\alpha$-blending to:
\begin{equation}
    \begin{split}
    \alpha_k &= \overline{\textrm{move}}(m_k, o_k) \prod_{j=1}^{k-1}\sqrt{(1- \overline{\textrm{move}}(m_j, o_j)^2)},
    \end{split}
\end{equation}
which ensures $\sum_{k=1}^K \alpha_k^2=1$ given an all-one background. For soft masks, we use two blobs with $c=0.05, s=20$ and $c=0.1, s=10$ respectively, where $s$ is a parameter that controls the blob size. We show results rendered by the modified $\alpha$-blending in \autoref{fig:blob}.

\section{Related Works}

\subsection{Text-to-image diffusion models}

Recently, diffusion models have demonstrated unprecedented results on text-to-image generation~\citep{ho2020denoising, nichol2021glide, dhariwal2021diffusion, rombach2021high, saharia2022photorealistic}, i.e., the task of generating an image from a textual description, by learning to progressively denoise an image from an input standard Gaussian noise. In the literature, T2I models vary with different design choices, including generation in pixel space~\citep{saharia2022photorealistic} or latent space~\citep{rombach2021high} and different denoiser architectures including U-Net~\citep{ronneberger2015u}-based~\citep{ho2020denoising} or transformer~\citep{vaswani2017attention}-based~\citep{peebles2022scalable}. Unlike previous image editing approaches that leverage attention cues~\citep{hertz2022prompt, tumanyan2023plug, chefer2023attend, epstein2023diffusion} or feature correspondence~\citep{mou2023dragondiffusion, tang2023emergent, shi2023dragdiffusion}, our approach is agnostic to the specific design choice of the denoiser.

\subsection{Layout conditioned image diffusion}
Extensive study has been made to add layout conditions to text-to-image diffusion. For training-free approaches, MultiDiffusion~\citep{bar2023multidiffusion} and locally conditioned diffusion~\citep{po2023compositional} predict noise using local prompts and composite them with region masking, Layout-Guidance~\citep{chen2023training} leverages the cross-attention map to provide the spatial guidance. For training-based approaches, ControlNet~\cite{zhang2023adding} and GLIGEN~\citep{li2023gligen} finetunes the pretrained image diffusion model on paired layout-image datasets. Different from the setting in this paper, they do not focus on spatial disentanglement, thus changing layouts will also affect contents. Additionally, a line of work studies joint layout and content conditioning. Paint-by-Example~\citep{yang2023paint} position reference objects to specific locations of a given image through additional model tuning, Collage Diffusion~\citep{sarukkai2023collage} harmonizes the collage of reference images using the image-to-image technique~\citep{meng2021sdedit} improved by ControlNet~\citep{zhang2023adding}. Recently, a concurrent work Anydoor~\citep{chen2023anydoor} demonstrates object moving using the paint-by-example pipeline. Our framework provides a mid-level representation and hence enables controllable scene generation, which is beyond the capability of these works.


\section{Experiment Details}
\subsection{Dataset}
\paragraph{Caption Generation.}
We use a large language model to automatically generate image captions. The prompt we used is:
\emph{Please give me 100 image captions that describe a single subject in a scene. The format is as follows: ``A cat is sitting in a museum. Subject: cat. Scene: museum.". ``Cat" is the subject and ``museum" is the scene.} Example image captions are as follows:
\begin{enumerate}[font=\itshape]
\item \emph{A bird is perched on a windowsill. Subject: bird. Scene: windowsill.}
\item \emph{A goldfish swims in a bowl. Subject: goldfish. Scene: bowl.}
\item  \emph{A kite soars above the beach. Subject: kite. Scene: beach.}
\item \emph{A bicycle leans against a brick wall. Subject: bicycle. Scene: brick wall.}
\item \emph{A turtle crawls along a sandy path. Subject: turtle. Scene: sandy path.}
\item \emph{A sunflower stands tall in a garden. Subject: sunflower. Scene: garden.}
\item \emph{A butterfly rests on a blooming flower. Subject: butterfly. Scene: blooming flower.}
\item \emph{A tree casts its shadow on a playground. Subject: tree. Scene: playground.}
\item \emph{A cloud drifts over a mountain peak. Subject: cloud. Scene: mountain peak.}
\item \emph{A snake slithers through the tall grass. Subject: snake. Scene: tall grass.}
\end{enumerate}
Subject and scene descriptions are used as foreground and background local descriptions respectively. We query the language models 10 times to collect 1,000 image captions.
\paragraph{Image Generation.}
We use an open-source $512 \times 512$ text-to-image latent diffusion model to generate images from the image captions. We generate 20 images for each caption, which results in 20,000 images. Then, we use an open-vocabulary segmentation model GroundedSAM~\citep{liu2023grounding} to segment the foreground object. The following rule-based filters are used to remove images with no or ambiguous foreground objects:
\begin{itemize}
    \item No bounding box detected.
    \item Bounding box confidence lower than 0.5.
    \item Bounding box area is larger than 60\% of the image size.
    \item Segmentation mask is smaller than 5\% of the image size.
\end{itemize}
5,092 images are left after filtering. Each image is associated with an image caption, local descriptions, and a segmentation mask.

\subsection{Metrics}
We detail evaluation metrics as follows:
\begin{itemize}
\item  \textbf{Mask IoU.} We employ the segmentation model to predict the foreground mask on the generated images. One of the two target layouts contains the original annotated mask. We can, therefore, compute a mask IoU between the annotated mask and the shifted mask. 
\item \textbf{Consistency.} We compute the mask IoU between the foreground masks for the two generated images. To compensate for masks that move out of the canvas, we align the masks in two different layouts respectively and take maximum IoU. 
\item \textbf{Visual Consistency.} For two images generated from different layouts, we segment foreground objects out, paste them on the same location on a white canvas, and compute LPIPS to measure object-level visual consistency.
\item \textbf{LPIPS.} We compute the LPIPS distance between the two generated views to examine the cross-view perceptual consistency. 
\item \textbf{SSIM.} We compute the SSIM similarity between the two generated views to examine the structural similarity.
\item \textbf{FID.} We compute the FID between the edited images and the test dataset to evaluate the image quality.
\end{itemize}
In addition, we report KID and CLIP Score.
\begin{itemize}
\item \textbf{KID.} Similar to FID, we report KID as well for image quality evaluation. 
\item \textbf{CLIP Score.} We measure the similarity between the image embedding and the text embedding to ensure that the text alignment does not degrade after editing.
\end{itemize}
\subsection{Implementation}
We implement our approach on the Diffusers library using publicly available text-to-image latent diffusion models. It employs a $64\times 64$ latent and generates $512\times 512$ image. For classifier-free guidance~\citep{ho2022classifier}, we set the guidance scale to 7.5. We employ the DDIM sampler~\citep{song2020denoising} and the number of sampling steps is 50. For most qualitative experiments, we set $N=8$, $\tau=25$, and $\mu_k, \nu_k$ to 40\% of the image size. For image editing experiments, we use GroundedSAM~\citep{liu2023grounding} to segment objects and use the segmentation masks as layer masks with manually assigned local prompts. We run all experiments on a single machine equipped with 8 32GB NVIDIA V100 GPUs. With multi-GPU parallelization, the total running time of a scene optimization and inference is less than 5 seconds.

\section{Qualitative Results}
\subsection{More generated scenes}
\begin{figure*}[h!]
    \centering
    \includegraphics[width=\textwidth]{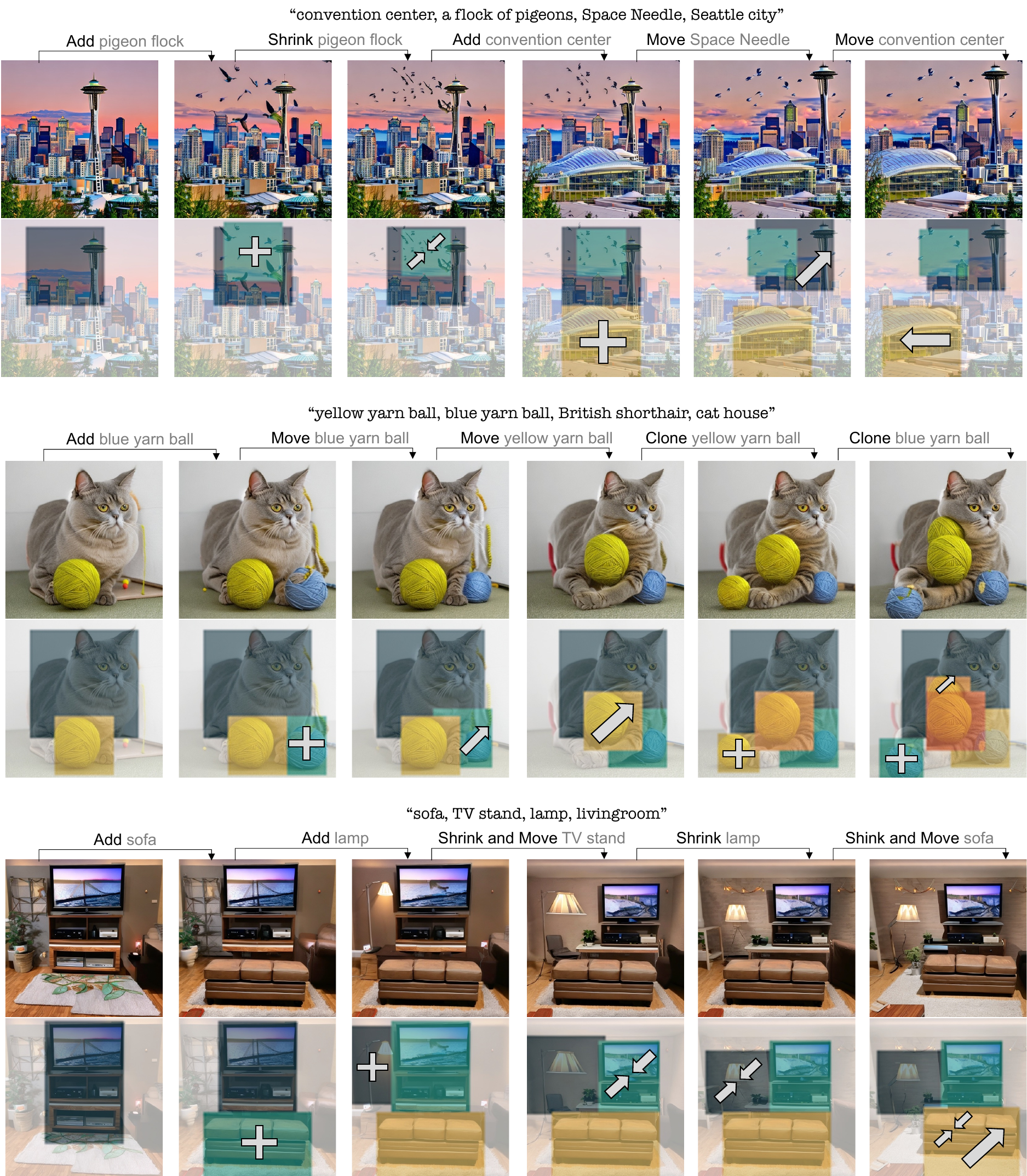}
    \caption{\textbf{More examples of generated controllable scene.} We apply sequential manipulations using the layered control.}
    \vspace{15pt}
    \label{fig:scene_more}
\end{figure*}
We show more examples of controllable scene generation in ~\autoref{fig:scene_more}.

\subsection{Comparison of object moving}
\begin{figure*}[t!]
    \centering
    \includegraphics[width=0.85\textwidth]{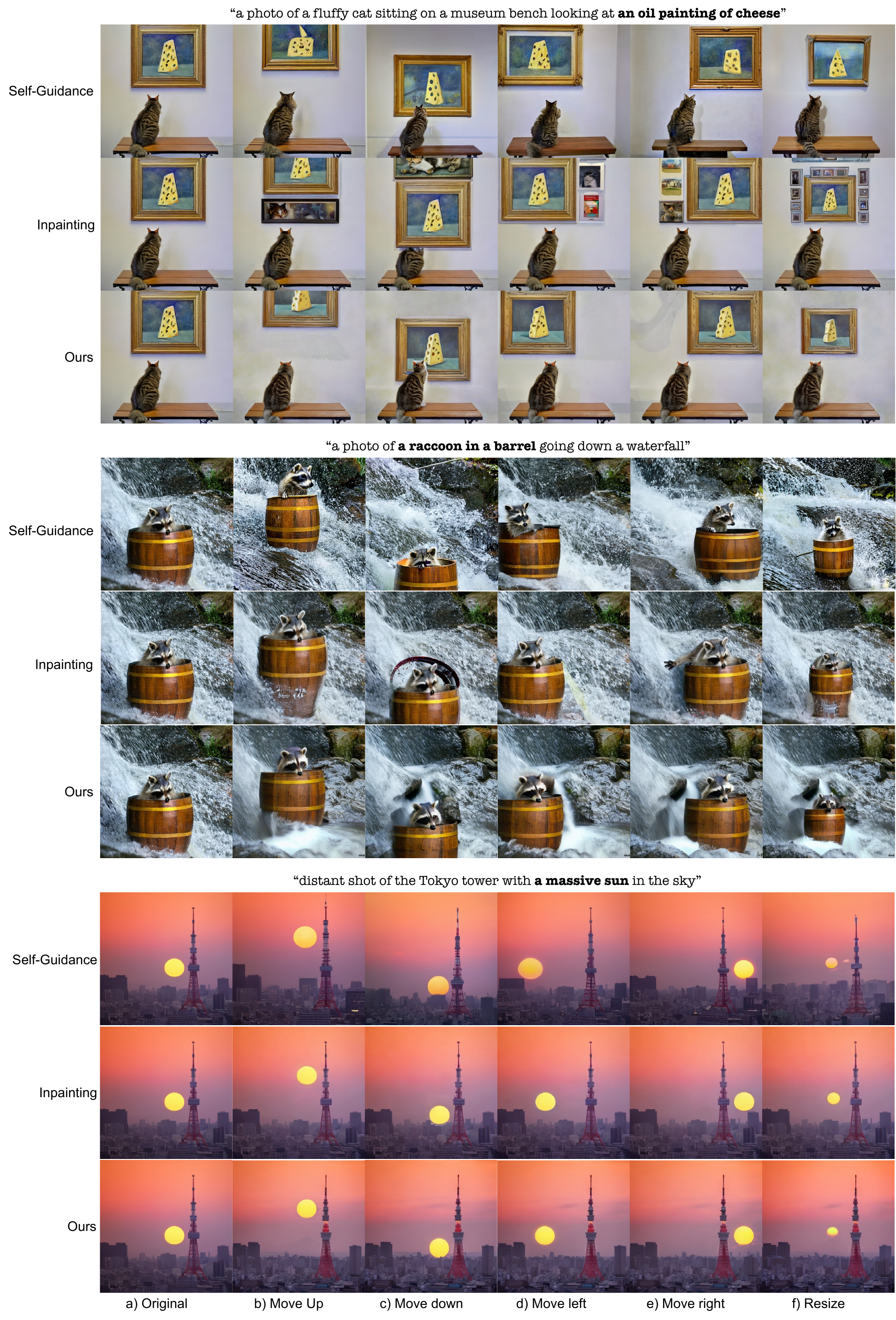}
    \caption{
    \textbf{Qualitative comparison on object moving.} Self-Guidance~\citep{epstein2023diffusion} and inpainting generates varing content across editings.
    }
    \label{fig:move_comparison_full}
\end{figure*}
We provide a comparison with Self-Guidance~\citep{epstein2023diffusion} and a specialized inpainting model on object moving in ~\autoref{fig:move_comparison_full}.

\subsection{Real image editing}
\begin{figure*}[t!]
    \centering
    \includegraphics[width=\textwidth]{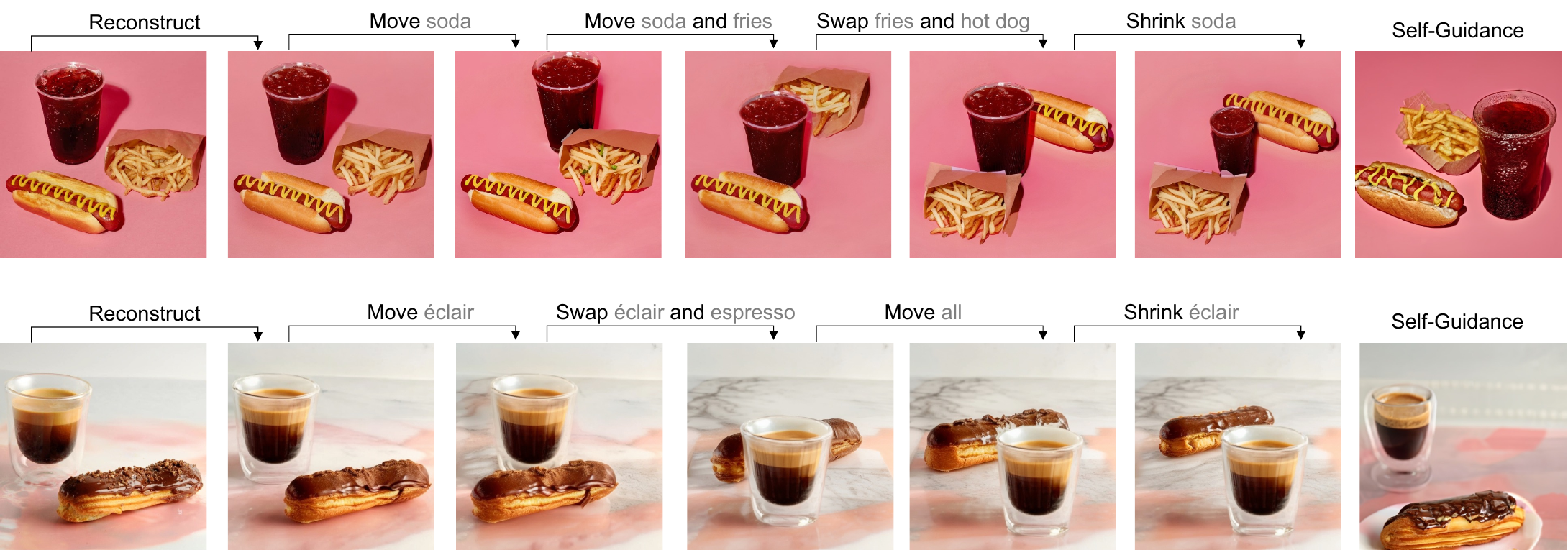}
    \caption{
    \textbf{Multi-object moving on real images.} Examples are borrowed from~\citet{epstein2023diffusion}.
    }
    \label{fig:real_image_editing}
\end{figure*}
Our approach can edit in-the-wild images. We demonstrate multi-object moving on real images using examples provided by~\citet{epstein2023diffusion} in \autoref{fig:real_image_editing}.

\subsection{Compatibility with different denoisers}
\begin{figure*}[h!]
    \centering
    \includegraphics[width=\textwidth]{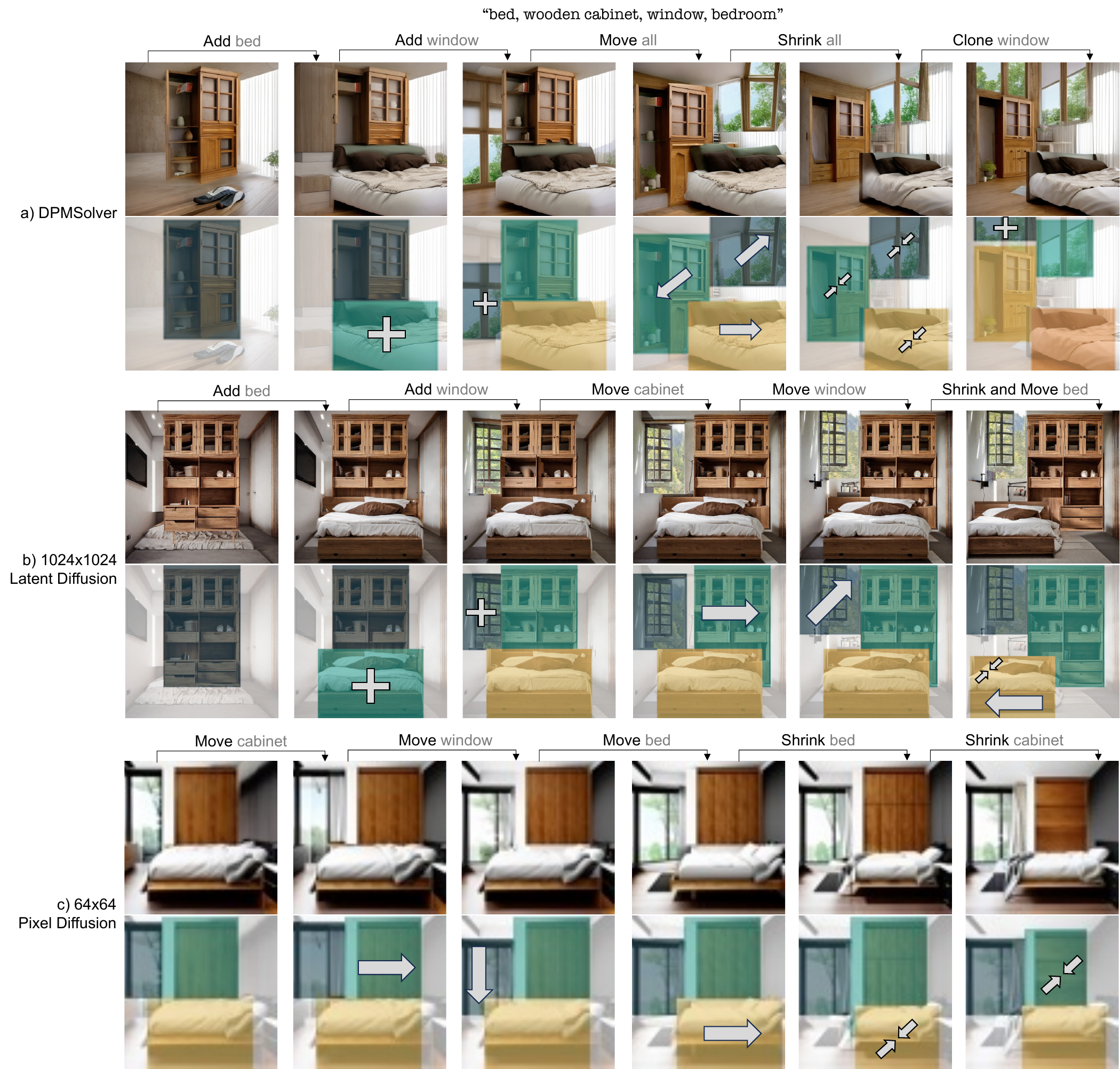}
    \caption{
    \textbf{Diffusion sampler and architecture.} We present editing results with different diffusion samplers and denoiser architectures to show our method is applicable in various configurations.
    }
    \label{fig:compatibility}
\end{figure*}
Our approach is compatible with general text-to-image diffusion models. We use a DDIM sampler and a $512\times512$ latent diffusion model in the main paper and show in ~\autoref{fig:compatibility} that our approach also works with different samplers:
\begin{itemize}
    \item \textbf{DPMSolver.} We set $T=25$ and $\tau=12$ and the inference gets even faster. We use the same random seed as the scene shown in Figure 1-Top to show the difference from DDIM-sampled results.
\end{itemize}
and different denoiser architectures:
\begin{itemize}
    \item \textbf{An open source $1024 \times 1024$ latent diffusion model.} The model has a larger latent space and generates higher-resolution images compared to the model we used in the main paper. It also employs a different language conditioning mechanism.
    \item \textbf{An open source pixel diffusion model.} The model denoises on the pixel space. It has three stages, the first stage generates a $64 \times 64$ image, and the second and the third stage upsample the image to $1024 \times 1024$ resolution. Here we only show the output from the first stage.
\end{itemize}

\subsection{Different random seeds}
\begin{figure*}[h!]
    \centering
    \includegraphics[width=\textwidth]{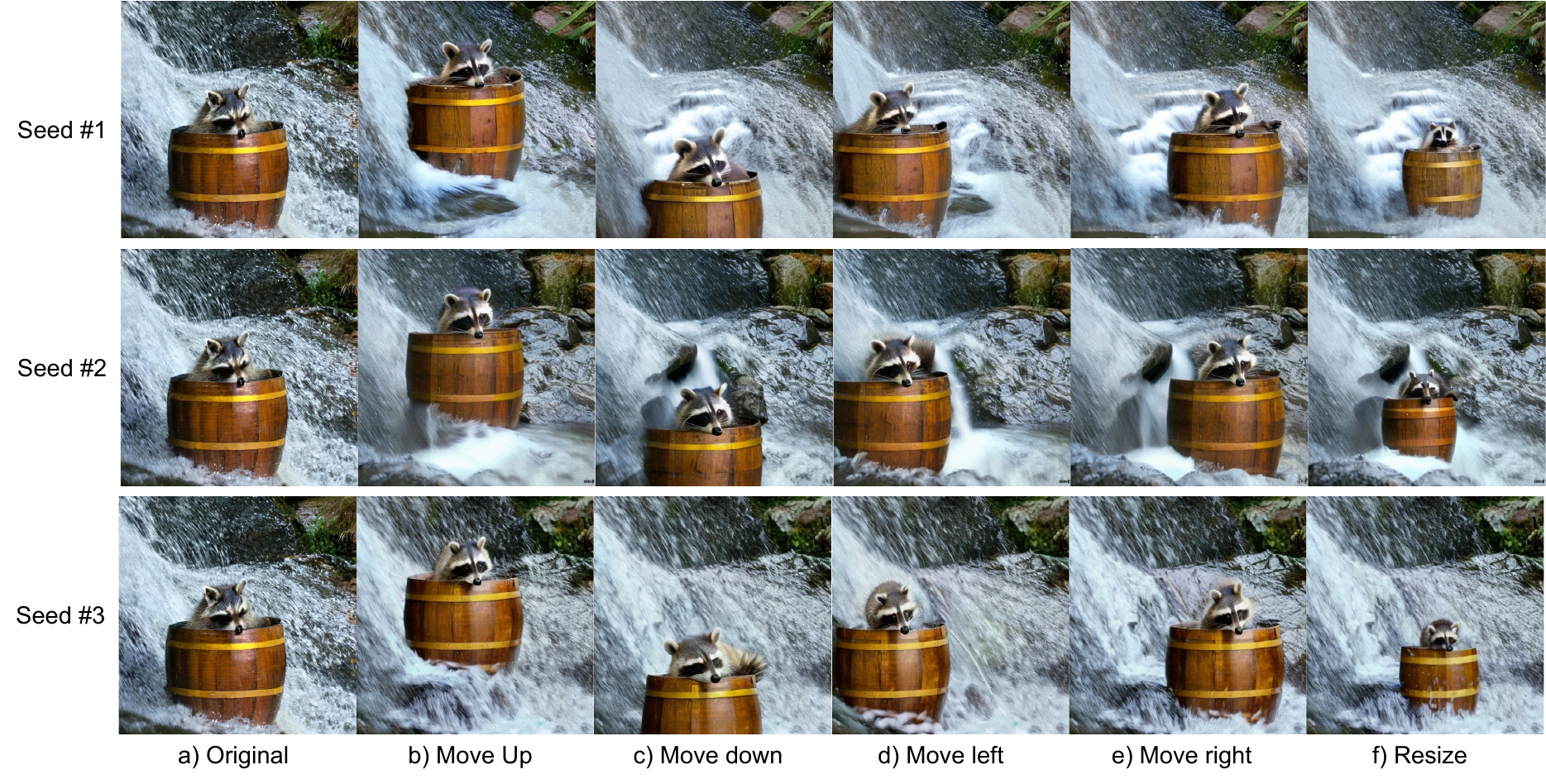}
    \caption{
    Results with different random seeds in the object moving task.
    }
    \label{fig:move_random_seeds}
\end{figure*}
Although our approach keeps the content consistent in different views of a scene, the randomness can be introduced by changing the random noise during initialization. We show the results of three different random seeds for the object moving tasks in~\autoref{fig:move_random_seeds}.

\subsection{Scenes after object replacement}
\begin{figure*}[t!]
    \centering
    \includegraphics[width=\textwidth]{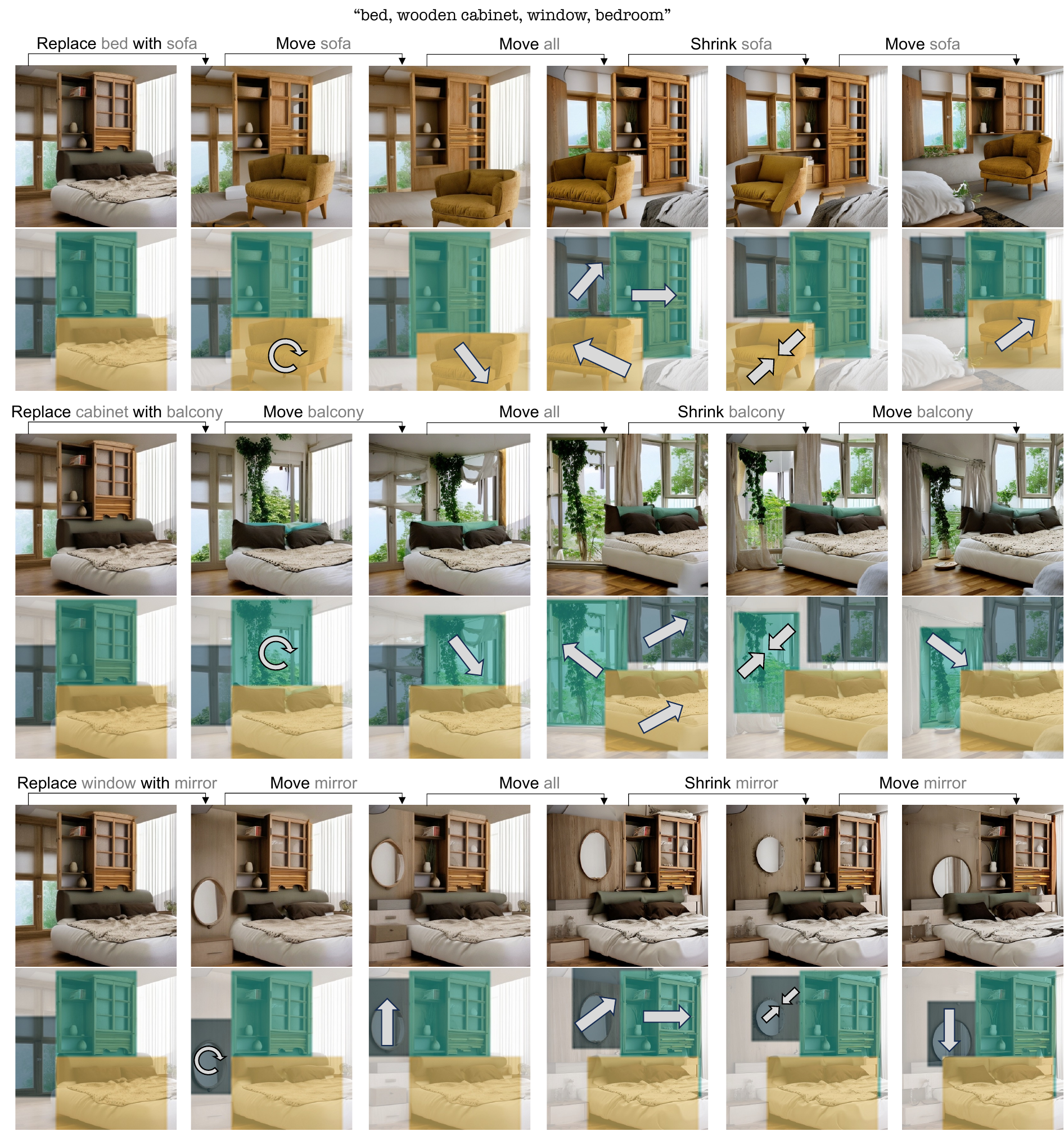}
    \caption{
    \textbf{Manipulating scenes with replaced objects.} We first replace an object in the scene before manipulating the scene layout show the corresponding editing results.
    }
    \label{fig:move_replace}
\end{figure*}
A scene remains rearrangeable after object replacement. We show results of manipulating scenes with replaced objects in~\autoref{fig:move_replace}.

\section{Quantitative Results}
\subsection{Full results for controllable scene generation}
\begin{table*}[h!]
\small
\caption{\textbf{Quantitative comparison for controllable scene generation.} $\dagger$: without the solid color bootstrapping strategy.}
\vspace{-10pt}
\begin{center}
\begin{tabular}{lcccc}
\toprule
Method & Mask IoU $\uparrow$ &  Consistency $\uparrow$ & LPIPS $\downarrow$ & SSIM $\uparrow$\\
\midrule
MultiDiffusion~\citep{bar2023multidiffusion}$^\dagger$ & 0.263  $\pm$ 0.004 & 0.257  $\pm$ 0.002 & 0.521  $\pm$ 0.002 & 0.450  $\pm$ 0.002 \\
MultiDiffusion~\citep{bar2023multidiffusion} & 0.466  $\pm$ 0.001 & 0.436  $\pm$ 0.004 & 0.519  $\pm$ 0.001 & 0.471 $\pm$ 0.002\\
\midrule
Ours$^\dagger$ & 0.310  $\pm$ 0.002 & 0.609  $\pm$ 0.003 & \bf{0.198}  $\pm$ 0.001 & 0.761  $\pm$ 0.001 \\
Ours &  \bf{0.522}  $\pm$ 0.001 &  \bf{0.721}  $\pm$ 0.002 & 0.215  $\pm$ 0.001 & \bf{0.762}  $\pm$ 0.000\\
\bottomrule
\end{tabular}
\label{tab:scene_comparison_full}
\end{center}
\end{table*}
We show full results for controllable scene generation with standard deviations in~\autoref{tab:scene_comparison_full}.

\subsection{Full results for object moving comparions}
\setlength{\tabcolsep}{4pt}
\begin{table*}[h!]
\caption{\textbf{Object moving comparison of RePaint~\citep{lugmayr2022repaint}, Inpainting, and our method.} $\dagger$: Inpainting means a specialized inpainting model trained with masking.}
\small
\begin{center}
\begin{tabular}{lcccccc}
\toprule
Method & FID $\downarrow$ & KID $_{\times 10^3}$ $\downarrow$ & Mask IOU $\uparrow$ &  CLIP Score $\uparrow$ & LPIPS $\downarrow$ & SSIM $\uparrow$\\
\midrule
RePaint  & 10.267 $\pm$ 0.020  & 1.167 $\pm$ 0.026 & 0.620 $\pm$ 0.001 & 0.321 $\pm$ 0.000 & 0.278 $\pm$ 0.001 & 0.671 $\pm$ 0.000\\
Inpainting$^\dagger$ & 6.383 $\pm$ 0.039 & 0.099 $\pm$ 0.014 & 0.747 $\pm$ 0.002 & 0.321 $\pm$ 0.000 & 0.264 $\pm$ 0.001 & 0.680 $\pm$ 0.001 \\
\midrule
Ours & \bf{5.289} $\pm$ 0.022 & \bf{0.059} $\pm$ 0.014 & \bf{0.817} $\pm$ 0.003 & 0.321 $\pm$ 0.000 & \bf{0.263} $\pm$ 0.001 & \bf{0.709} $\pm$ 0.000 \\
\bottomrule
\end{tabular}
\label{tab:move_comparison_full}
\end{center}
\end{table*}
We present full results for object moving comparisons with standard deviations, KID, and CLIP score in~\autoref{tab:move_comparison_full}

\subsection{Full results for ablation on scene generation}
\begin{table*}[h!]
\small
\caption{\textbf{Ablation on controllable scene generation.} We compare our method by varying the number of views $N$ and image diffusion steps $\tau$. $\dagger$: Layout using deterministic sampling at fixed intervals.}
\begin{center}
\begin{tabular}{lccccc}
\toprule
$N$ & $\tau$ & Mask IoU $\uparrow$ &  Consistency $\uparrow$ & LPIPS $\downarrow$ & SSIM $\uparrow$\\
\midrule
2 & 25 &  0.477 $\pm$ 0.020 & 0.619 $\pm$ 0.017 & 0.274 $\pm$ 0.004 & 0.697 $\pm$ 0.004 \\
8$^\dagger$ & 25&  0.485 $\pm$ 0.006 & 0.638 $\pm$ 0.011 & 0.269 $\pm$ 0.002 & 0.699 $\pm$ 0.004\\
8 & 25&  0.499 $\pm$ 0.005 & 0.657 $\pm$ 0.012 & 0.274 $\pm$ 0.001 & 0.689 $\pm$ 0.004\\
\midrule
2 & 25 & 0.477 $\pm$ 0.020 & 0.619 $\pm$ 0.017 & 0.274 $\pm$ 0.004 & 0.697 $\pm$ 0.004 \\
2 & 13 & 0.483 $\pm$ 0.024 & 0.661 $\pm$ 0.023 & 0.227 $\pm$ 0.004 & 0.753 $\pm$ 0.003\\
2 & 0 &  0.501 $\pm$ 0.015 & 0.699 $\pm$ 0.019 & \bf{0.208} $\pm$ 0.005 & \bf{0.778} $\pm$ 0.004\\
\midrule
8 & 0 &  \bf{0.515} $\pm$ 0.010 &  \bf{0.723} $\pm$ 0.016 & 0.211 $\pm$ 0.002 & 0.767 $\pm$ 0.003\\
\bottomrule
\end{tabular}
\label{tab:scene_ablation_full}
\end{center}
\end{table*}
We show full results for $N$ and $\tau$ ablation on controllable scene generation with standard deviations in~\autoref{tab:scene_ablation_full}.

\subsection{Additional results for object moving ablation}
\begin{table*}[h!]
\caption{\textbf{Object moving ablation.} We compare our method with inpainting-based approaches on object moving for varying number of views $N$ and image diffusion steps $\tau$.}
\small
\begin{center}
\begin{tabular}{lccccccc}
\toprule
$N$ & $\tau$ & FID $\downarrow$ & KID $\downarrow$ & Mask IOU $\uparrow$ &  CLIP Score $\uparrow$ & LPIPS $\downarrow$ & SSIM $\uparrow$\\
\midrule
2 & 25 & 5.918 $\pm$ 0.018 & -0.020 $\pm$ 0.004 & 0.788 $\pm$ 0.003 & 0.322 $\pm$ 0.000 & 0.294 $\pm$ 0.001 & 0.672 $\pm$ 0.001 \\
8 & 25 &  5.890 $\pm$ 0.032 & -0.010 $\pm$ 0.004 & 0.794 $\pm$ 0.002 & 0.321 $\pm$ 0.000 & 0.289 $\pm$ 0.001 & 0.676 $\pm$ 0.000\\
\midrule
2 & 38 & 7.401 $\pm$ 0.025 & \bf{-0.079} $\pm$ 0.009 & 0.667 $\pm$ 0.003 & 0.322 $\pm$ 0.000 & 0.368 $\pm$ 0.001 & 0.598 $\pm$ 0.001\\
2 & 25 & 5.918 $\pm$ 0.018 & -0.020 $\pm$ 0.004 & 0.788 $\pm$ 0.003 & 0.322 $\pm$ 0.000 & 0.294 $\pm$ 0.001 & 0.672 $\pm$ 0.001 \\
2 & 13 &  \bf{5.289} $\pm$ 0.022 & 0.059 $\pm$ 0.014 & 0.817 $\pm$ 0.003 & 0.321 $\pm$ 0.000 & 0.263 $\pm$ 0.001 & 0.709 $\pm$ 0.000 \\
2 & 0 &  5.320 $\pm$ 0.029 & 0.182 $\pm$ 0.020 & \bf{0.836} $\pm$ 0.003 & 0.322 $\pm$ 0.000 & \bf{0.255}$\pm$ 0.001 & \bf{0.722} $\pm$ 0.001\\
\bottomrule
\end{tabular}
\label{tab:move_ablation_full}
\end{center}
\end{table*}
We provide additional results for $N$ and $\tau$ ablation on object moving in~\autoref{tab:move_ablation_full}.

\clearpage
\end{document}